\documentclass[10pt,twocolumn,letterpaper]{article}

\usepackage[pagenumbers]{cvpr} % To force page numbers, e.g. for an arXiv version

\definecolor{cvprblue}{rgb}{0.21,0.49,0.74}
\usepackage[pagebackref,breaklinks,colorlinks,allcolors=cvprblue]{hyperref}
\newcommand{\benchmark}{{\sc DispBench}}
\usepackage{booktabs}
\usepackage{algpseudocode}
\usepackage{algorithm}
\usepackage{multirow}
\usepackage{multicol}
\usepackage{mathtools}
\usepackage{xcolor}
\usepackage{ragged2e}
\usepackage{graphicx}
\usepackage[export]{adjustbox}
\usepackage{tikz}
\usepackage[frozencache=true,cachedir=_minted-output]{minted}

\usepackage{soul}

\usepackage{newfloat}
\usepackage{listings}

\newcommand\blfootnote[1]{%
  \begingroup
  \renewcommand\thefootnote{}\footnote{#1}%
  \addtocounter{footnote}{-1}%
  \endgroup
}

\usepackage[capitalize]{cleveref}
\crefname{section}{Sec.}{Secs.}
\Crefname{section}{Section}{Sections}
\Crefname{table}{Table}{Tables}
\crefname{table}{Tab.}{Tabs.}
\definecolor{cadmiumgreen}{rgb}{0.0, 0.42, 0.24}
\definecolor{custom}{cmyk}{0.1,0.48,0.49,0.2}
\definecolor{OliveGreen}{cmyk}{0.64,0,0.95,0.40}
\definecolor{new}{rgb}{0.81,0.05,0.9}
\definecolor{BrickRed}{rgb}{0.81,0.1,0.1}
\definecolor{RoyalBlue}{rgb}{0.2,0.2,0.75}

   \usepackage{amsmath,amssymb}
   \usepackage{bm}
\makeatletter
\DeclareRobustCommand\onedot{\futurelet\@let@token\@onedot}
\def\@onedot{\ifx\@let@token.\else.\null\fi\xspace}

\makeatother

\def\clap#1{\hbox to 0pt{\hss #1\hss}}%
\def\initials#1{\protect\clap{\protect\smash{\protect\raisebox{1.4ex}{\protect\tiny{\protect\textsf{\protect\textit{#1}}}}}}}%
\makeatletter
\newcommand{\EDIT}[4][]{\protect\@ifundefined{hidecomments}{%
  \protect\strut{\color{#3}{\hspace{0pt}\initials{#2}\protect\sout{#1}{~#4}}}%
  }{#4}}
\newcommand{\NOTEboxed}[3]{\protect\@ifundefined{hidecomments}{%
  {\begin{center}\fbox{\parbox{0.97\linewidth}{\protect\EDIT{#1}{#2}{#3}}}\end{center}}
  }{}}
\newcommand{\COMM}[3]{\protect\@ifundefined{hidecomments}{%
  {\protect\EDIT{#1}{#2}{#3}}%
  }{}}
\newcommand{\DefAuthor}[2] % initials, color
{%
  \expandafter\newcommand\csname #1edit\endcsname[2][]{\protect\EDIT[##1]{#1}{#2}{##2}}
  \expandafter\newcommand\csname #1\endcsname[1]{\protect\COMM{#1}{#2}{##1}}
  \expandafter\newcommand\csname #1boxed\endcsname[1]{\protect\NOTEboxed{#1}{#2}{##1}}
}
\definecolor{dkgreen}       {rgb}{0.0,0.5,0.0}
\definecolor{dkblue}        {rgb}{0.0,0.0,0.7}
\definecolor{dkcyan}        {rgb}{0.0,0.5,0.5}
\definecolor{dkmagenta}     {rgb}{0.5,0.0,0.5}
\DefAuthor{MK}{dkmagenta} % MARGRET'S COMMENTS
\DefAuthor{AB}{dkgreen} % ANDRES' COMMENTS
\DefAuthor{JS}{dkblue} % JENNY'S COMMENTS
\DefAuthor{SC}{dkcyan} % STUDENTS' COMMENTS
\DefAuthor{SA}{orange} % SHASHANK'S COMMENTS
\DefAuthor{todo}{red}
\usepackage{pifont}% http://ctan.org/pkg/pifont
\def\paperID{0006} % *** Enter the Paper ID here
\def\confName{CVPR}
\def\confYear{2025}

\title{\benchmark: Benchmarking Disparity Estimation to Synthetic Corruptions}

\author{Shashank Agnihotri$^{*,1}$
\and
Amaan Ansari$^{*,1}$
\and
Annika Dackermann$^{*,1}$
\and
Fabian Rösch$^{*,1}$
\and
Margret Keuper$^{1,2}$\\
$^{1}$Data and Web Science Group, University of Mannheim, Germany \\
$^{2}$Max-Planck-Institute for Informatics, Saarland Informatics Campus, Germany \\
{\tt\small shashank.agnihotri@uni-mannheim.de}
}

\begin{document}
\maketitle
\begin{abstract}
Deep learning (DL) has surpassed human performance on standard benchmarks, driving its widespread adoption in computer vision tasks. One such task is disparity estimation, estimating the disparity between matching pixels in stereo image pairs, which is crucial for safety-critical applications like medical surgeries and autonomous navigation. However, DL-based disparity estimation methods are highly susceptible to distribution shifts and adversarial attacks, raising concerns about their reliability and generalization. Despite these concerns, a standardized benchmark for evaluating the robustness of disparity estimation methods remains absent, hindering progress in the field.

To address this gap, we introduce \benchmark{}, a comprehensive benchmarking tool for systematically assessing the reliability of disparity estimation methods. \benchmark{} evaluates robustness against synthetic image corruptions such as adversarial attacks and out-of-distribution shifts caused by 2D Common Corruptions across multiple datasets and diverse corruption scenarios. We conduct the most extensive performance and robustness analysis of disparity estimation methods to date, uncovering key correlations between accuracy, reliability, and generalization. \href{https://github.com/shashankskagnihotri/benchmarking_robustness/tree/disparity_estimation/final/disparity_estimation}{Open-source code for \benchmark{}}.
\end{abstract}
\blfootnote{Accepted at CVPR 2025 Workshop on Synthetic Data for Computer Vision. $^{*}$Equal Contribution.}

\section{Background}
\label{sec:intro}
%\SA{Shashank's comments}.
%\SC{Students' comments}.
%\JS{Jenny's comments}.
%\AB{Andres' comments}.
%\MK{Margret's comments}.

\begin{figure}[t]
    \centering
    \scalebox{1.0}{
    \begin{tabular}{c}
    \includegraphics[width=\linewidth]{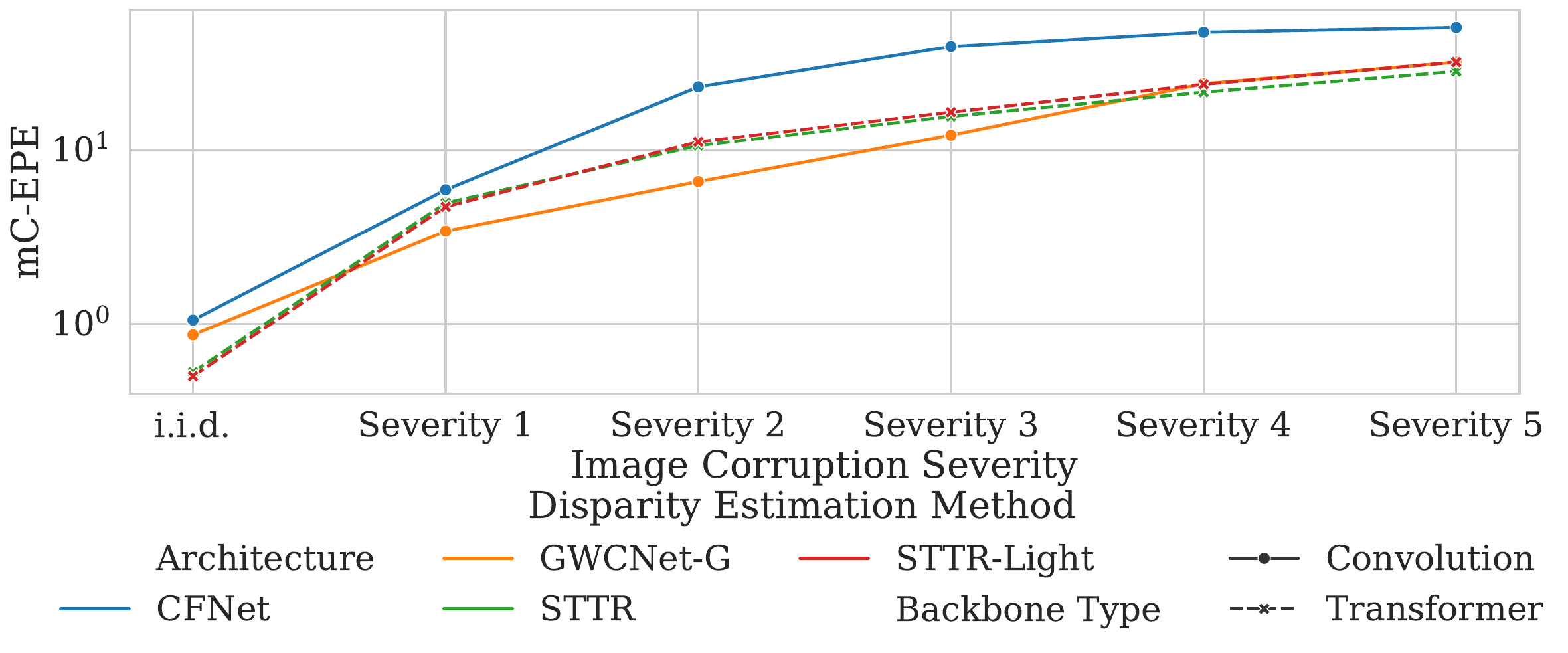}
         \\
    \end{tabular}
    }
    \caption{Analyzing the generalization ability of some Disparity estimation methods: GWCNet~\cite{gwcnet}, CFNet~\cite{cfnet}, and STTR and STTR-light~\cite{sttr} proposed over time. The y-axis represents the mean End-Point-Error (EPE) on Syntheticc Corruptions (2D Common Corrruptions~\cite{commoncorruptions}) at different severalties (severity=0 is i.i.d. performance) using the FlyingThings3D~\cite{flyingthings_dispnet}, i.e., lower is better. % and a higher error is worse. 
    We observe that disparity estimation methods lack the generalization ability to common corruptions and, thus, are not safe for real-world deployment.}
    \label{fig:teaser}
\end{figure}
The vision task of disparity estimation, also commonly known as stereo-matching is used to estimate the disparity between matching pixels in stereo image pairs.
\citet{flyingthings_dispnet} proposed the first Deep Learning (DL) based method for disparity estimation called DispNet.
This led to disparity estimation becoming primarily a DL-based task~\cite {sttr,gwcnet,cfnet,agnihotri2023improving}. 
However, DL-based methods are known to be unreliable~\citep{geirhos2018imagenet,prasad2022towards,agnihotri2023unreasonable,agnihotri2024beware,das20212}, they tend to learn shortcuts rather than meaningful feature representations~\citep{shortcut} and can be easily deteriorated even by small perturbations, causing the evaluation samples to not be independent and identically distributed (i.i.d.) w.r.t.~the training samples.
This shift from i.i.d. samples can be caused due to changes in the environment, changes in weather conditions, or image corruption due to sensor noise~\cite{commoncorruptions,sommerhoff2023differentiable,sommerhoff2024task,agnihotri2024roll,grabinski2022aliasing,yue2024improving,das2023weakly}. 
Such shifts cause the evaluations to be Out-Of-Distribution (OOD), and robustness to such shifts is called OOD Robustness.
OOD Robustness is often used as a metric for the generalization ability of a method~\cite{hoffmann2021towards,hendrycks2020augmix,grabinski2022robust,li2023intra,grabinski2022frequencylowcut}.
Another possible cause of distribution shifts could be either accidental or malicious adversarial attacks~\cite{li2024adversarial, medi2024towards,medi2025fair,schmalfuss2022perturbationconstrained,schmalfuss2022advsnow,scheurer2023detection}.
Here, the perturbations made to an image are optimized to fool the method while the semantic meaning of the images remains the same for a human observer.
When adversarial attacks are optimized with full information about a model and its loss, they are called white-box adversarial attacks.
Since these white-box attacks can potentially simulate the worst-case scenario for a method, they are often used as a proxy to measuring their reliability~\cite{agnihotri2023cospgd,pgd,fgsm}.

In \cref{fig:teaser}, we provide an overview of the i.i.d. performance, generalization ability, and reliability of disparity estimation methods proposed over time on the FlyingThings3D dataset~\cite{flyingthings_dispnet}.
We include old popular methods such as GWCNet and CFNet and new large transformer-based STTR and its lightweight version STTR-light, which, due to its training regime, are proposed as zero-shot disparity estimation methods. 
Here, we observe a disturbing pattern: while the i.i.d. performance has improved over time, since this improvement has been the focus of most works, the models still lack robustness.
This is particularly concerning as disparity estimation is often used in the real world, especially for safety-critical scenarios such as medical surgery procedures~\cite{disparity_surgery,yang2024disparity_surgery}, including invasive surgeries such as laparoscopy~\cite{muller2022fast_disparity_laproscopy_surgery} and in autonomous driving~\cite{chuah2021deep_disparity_driving}.
Here, safety is paramount, and to ensure the safe deployment of recent DL-based disparity estimation methods, their reliability and generalization ability need to be guaranteed. 
However, no such guarantees can be provided currently since no works focus on OOD and the adversarial robustness of disparity estimation methods.
This is primarily due to a lack of datasets that enable such studies.
Capturing corruptions in the wild and then annotating for disparity estimation is a time and resource intensive process.

Some prior works have focused on other kinds of robustness; for example, a recent work \cite{zhang2024robust_domain_shift_disparity} looks into the robustness of disparity estimation works to domain shifts, while \cite{reflection_robust_stereo,occlusion_robust_stereo} studies the robustness of methods to occlusions.
Currently, there exists no unified framework to evaluate disparity estimation methods for safe deployment in the real world.
\citet{OpenStereo} recently proposed a benchmarking tool for disparity estimation methods.
However, this tool is limited to i.i.d. performance evaluations.
This is a significant limitation impeding the community's ability to ensure safe, reliable, and generalizable DL-based disparity estimation methods for the real world.

To bridge this gap, we propose \benchmark{}, the first robustness benchmarking tool for disparity estimation.
\benchmark{} is easy to use and extending it to future disparity estimation methods and datasets, when they are proposed, is straightforward.
It is inspired by similar popular benchmarks for the image classification tasks~\cite{robustbench,tang2021robustart} and object detection~\cite{gupta2024robust_challenging_weather,michaelis2019benchmarking,adv_robust_obj_det_with_training,adv_robust_obj_det,extreme_construction_obj_det}.
It enables i.i.d. evaluations of various DL-based disparity estimation methods across multiple commonly used disparity estimation datasets.
It also facilitates research in the reliability and generalization ability of disparity estimation methods, as it enables users to use synthetic image corruptions, specifically, 5 diverse adversarial attacks and 15 established common corruptions.
This will help researchers build better models that are not limited to improved performance on identical and independently distributed (i.i.d.) samples and are less vulnerable to adversarial attacks while generalizing better to image corruptions.
Our proposed \benchmark{} facilitates this, streamlining it for future research to utilize.

The main contributions of this work are as follows:
\begin{itemize}
    \item We provide a benchmarking tool \benchmark{} to evaluate the performance of most DL-based disparity estimation methods over 2 different datasets and synthetic corruptions.
    %\item We provide a publicly available one-stop location for multiple checkpoints of different disparity estimation methods over different datasets, streamlining benchmarking while enabling the research community to add further checkpoints, methods, and datasets.
    \item We benchmark the aforementioned models against commonly used adversarial attacks and common corruptions that can be easily queried using \benchmark{}.
    \item We perform an in-depth analysis using \benchmark{} and present interesting findings showing methods that perform well on i.i.d.~are remarkably less reliable and generalize worse than other non-well-performing methods.
    \item We show that synthetic corruptions on synthetic datasets do not represent real-world corruptions; thus, synthetic corruptions on real-world datasets are required.
    %\item We analyze correlations between performance, reliability, and generalization abilities of disparity estimation methods, under various lenses such as point matching methods used, and the number of learnable parameters.    
\end{itemize}
%%%%%%%%%%%%%%%%%%%%%%%%%%%%%    END INTRODUCTION     %%%%%%%%%%%%%%%%%%%%%
%_________________________    START RELATED WORK    ____________________
%%%%%%%%%%%%%%%%%%%%%%%%%%%%%    END RELATED WORK     %%%%%%%%%%%%%%%%%%%%%
%_________________________    START METHODS    ____________________

\section{\benchmark{} Usage}
There exists no standardized tool for evaluating the performance of disparity estimation methods.
Thus, the codebase for such a tool had to be written from scratch.
In the following, we describe the benchmarking tool, \benchmark{}.
Currently, it supports 4 unique architectures (new architectures to be added to \benchmark{} with time) and 2 distinct datasets, namely FlyingThings3D~\citep{flyingthings_dispnet} and KITTI2015~\citep{kitti15} (please refer \cref{sec:appendix:dataset_details} for additional details on the datasets).
It enables training and evaluations on all aforementioned datasets, including evaluations using SotA adversarial attacks such as CosPGD~\citep{agnihotri2023cospgd} and other commonly used adversarial attacks like BIM~\citep{bim}, PGD~\citep{pgd}, FGSM~\citep{fgsm}, under various Lipshitz ($l_p$) norm bounds and APGD~\cite{apgd} under the $\ell_{\infty}$-norm bound. 
Additionally, it enables evaluations for Out-of-Distribution (OOD) robustness by corrupting the inference samples using 2D Common Corruptions~\citep{commoncorruptions}.

We follow the nomenclature set by RobustBench~\citep{robustbench} and use ``threat\_model'' to define the kind of evaluation to be performed.
When ``threat\_model'' is defined to be ``None'', the evaluation is performed on unperturbed and unaltered images, if the ``threat\_model'' is defined to be an adversarial attack, for example ``PGD'', ``CosPGD'' or ``BIM'', then \benchmark{} performs an adversarial attack using the user-defined parameters.
Whereas, if ``threat\_model'' is defined to be ``2DCommonCorruptions'', the \benchmark{} performs evaluations after perturbing the images with 2D Common Corruptions.
If the queried evaluation already exists in the benchmark provided by this work, then \benchmark{} simply retrieves the evaluations, thus saving computation.
Please refer to \cref{sec:appendix:description} for details on usage. 

Following, we show the basic commands to use \benchmark{}. 
We describe each attack and common corruption supported by \benchmark{} in detail in \cref{sec:appendix:description}.
Please refer to \cref{sec:appendix:evaluation_details} for details on the arguments.

\subsection{Model Zoo}
\label{sec:usage:model_zoo}
It is challenging to find all checkpoints, whereas training them is time and compute-exhaustive.
Thus, we gather available model checkpoints made available online by the respective authors.
The trained checkpoints for all models available in \benchmark{} can be obtained using the following lines of code:
\begin{minted}[fontsize=\small, breaklines]{python}
from dispbench.evals import load_model
model = load_model(model_name='STTR', 
    dataset='KITTI2015')
\end{minted}
Each model checkpoint can be retrieved with the pair of `model\_name', the name of the model, and `dataset', the dataset for which the checkpoint was last fine-tuned.

\subsection{Adversarial Attacks}
\label{subsec:evaluation_details:adv_attack}
To evaluate a model for a given dataset on an attack, the following lines of code are required.
\begin{minted}[fontsize=\small, breaklines]{python}
from dispcbench.evals import evaluate
model, results = evaluate( 
 model_name='STTR', dataset='KITTI2015' retrieve_existing=True,
 threat_config='config.yml')
\end{minted}
Here, the `config.yml' contains the configuration for the threat model, for example, when the threat model is a PGD attack, `config.yml' could contain `threat\_model=\textit{``PGD''}', `iterations=\textit{20}', `alpha=\textit{0.01}', `epsilon=\textit{8}', and `lp\_norm=\textit{``Linf''}'.
The argument description is as follows:
\begin{itemize}    
\item `model\_name' is the name of the disparity estimation method to be used, given as a string.
\item `dataset' is the name of the dataset to be used also given as a string. 
\item `retrieve\_existing' is a boolean flag, which when set to `True' will retrieve the evaluation from the benchmark if the queried evaluation exists in the benchmark provided by this work, else \benchmark{} will perform the evaluation.
If the `retrieve\_existing' boolean flag is set to `False' then \benchmark{} will perform the evaluation even if the queried evaluation exists in the provided benchmark.
\item The `config.yml' contains the following:
\begin{itemize}
    \item `threat\_model' is the name of the adversarial attack to be used, given as a string.
    \item `iterations' are the number of attack iterations, given as an integer.
    \item `epsilon' is the permissible perturbation budget $\epsilon$ given a floating point (float).
    \item `alpha' is the step size of the attack, $\alpha$, given as a floating point (float).
    \item `lp\_norm' is the Lipschitz continuity norm ($l_p$-norm) to be used for bounding the perturbation, possible options are `Linf' and `L2' given as a string.
    \item `target' is false by default, but to do targeted attacks, either the user can set `target'=True, to use the default target of $\overrightarrow{0}$, or can pass a specific tensor to be used as the target. 
\end{itemize}
\end{itemize}
\begin{figure}
    \centering

    \scalebox{0.9}{
    \begin{tabular}{cc}
         \textbf{Left Image} & \textbf{Right Image}  \\

         \includegraphics[width=0.25\textwidth]{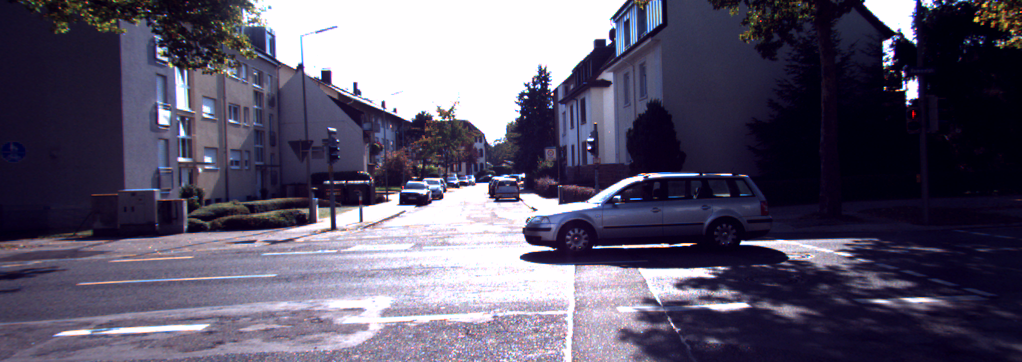}     
         &
         \includegraphics[width=0.25\textwidth]{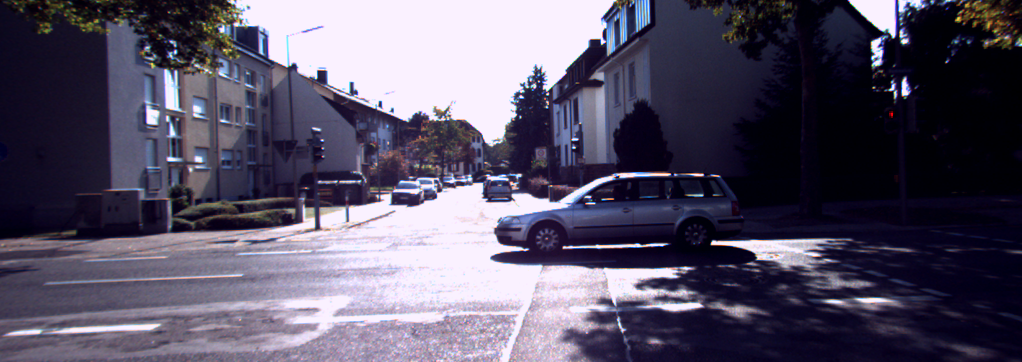}
         \\
         \midrule
         \multicolumn{2}{c}{Predictions} \\
         \midrule

         \multicolumn{2}{c}{i.i.d.~Performance} \\
         \multicolumn{2}{c}{\includegraphics[width=0.5\textwidth]{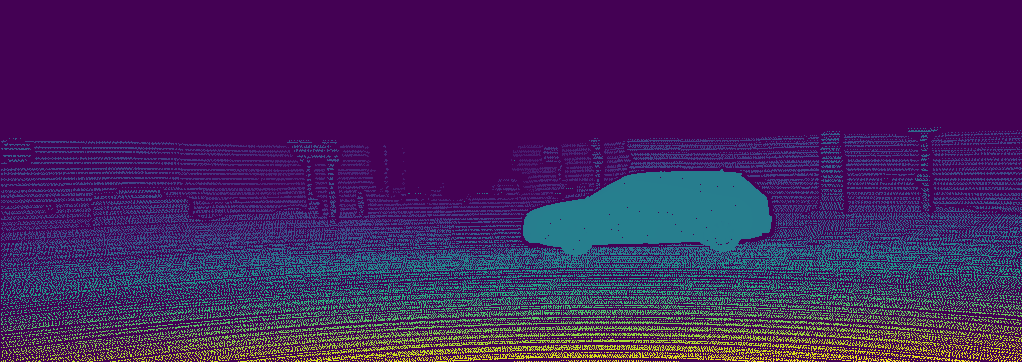}} \\

         \multicolumn{2}{c}{After FGSM attack} \\
         \multicolumn{2}{c}{\includegraphics[width=0.5\textwidth]{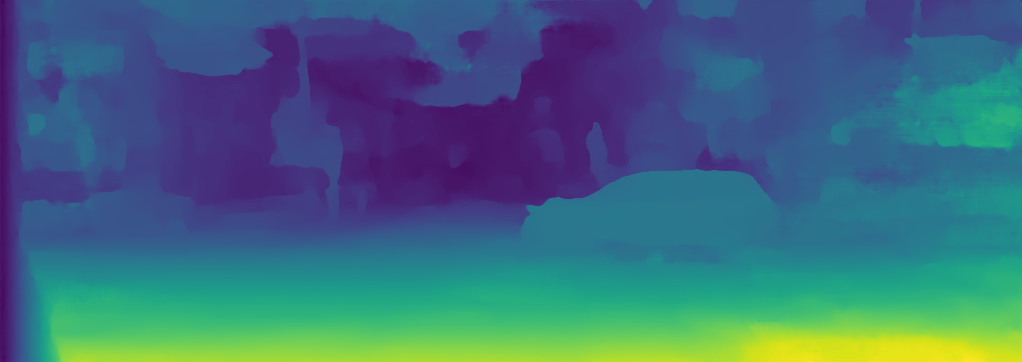}} \\

         \multicolumn{2}{c}{After 20 iteration BIM attack} \\
         \multicolumn{2}{c}{\includegraphics[width=0.5\textwidth]{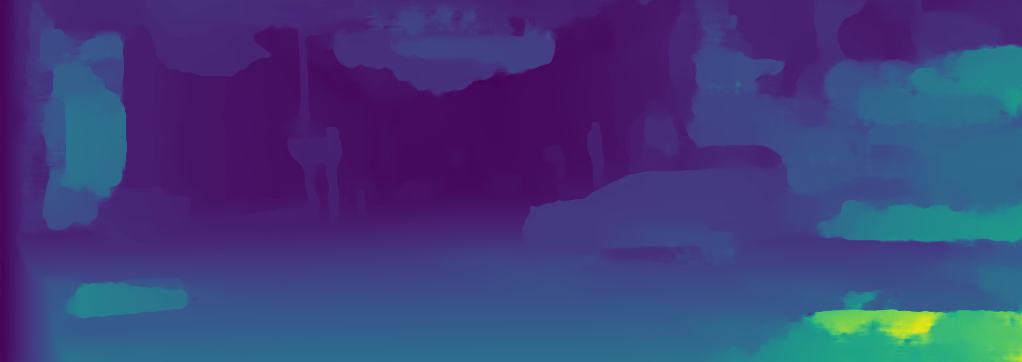}} \\

         \multicolumn{2}{c}{After 20 iteration PGD attack} \\
         \multicolumn{2}{c}{\includegraphics[width=0.5\textwidth]{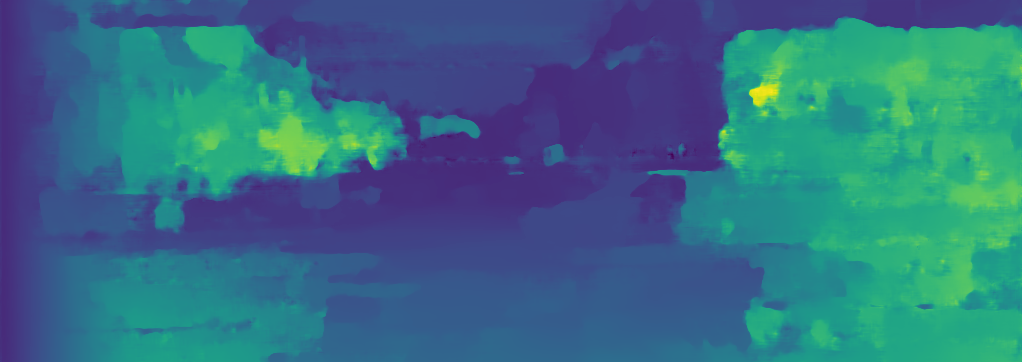}} \\

         \multicolumn{2}{c}{After 20 iteration CosPGD attack} \\
         \multicolumn{2}{c}{\includegraphics[width=0.5\textwidth]{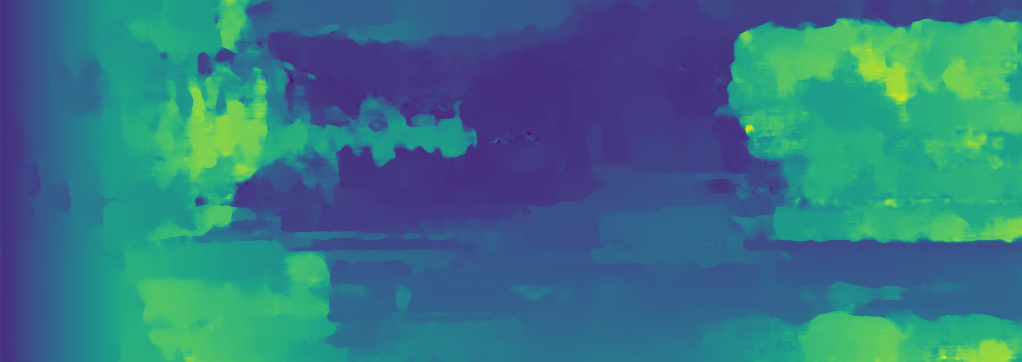}} \\

    \end{tabular}
    }
    \caption{Example of performing adversarial attacks on STTR using KITTI2015 dataset under different attacks. We show the samples before and after the attacks and the predictions before and after the respective adversarial attacks.}
    \label{fig:adv_exp_kitti_sttr}
\end{figure}

\begin{figure*}
    \centering

    \scalebox{1.0}{
    \begin{tabular}{@{}c@{\,\,}c@{\,\,}c@{}}
         \textbf{Left Image} & \textbf{Right Image} & \textbf{Zoomed-in Prediction} \\

         \multicolumn{3}{c}{i.i.d.~Performance} \\
         \begin{tikzpicture}
        \node[anchor=south west, inner sep=0] (image) at (0,0) {\includegraphics[clip,trim=0cm 0cm 0cm 3.425cm, width=0.32\linewidth]{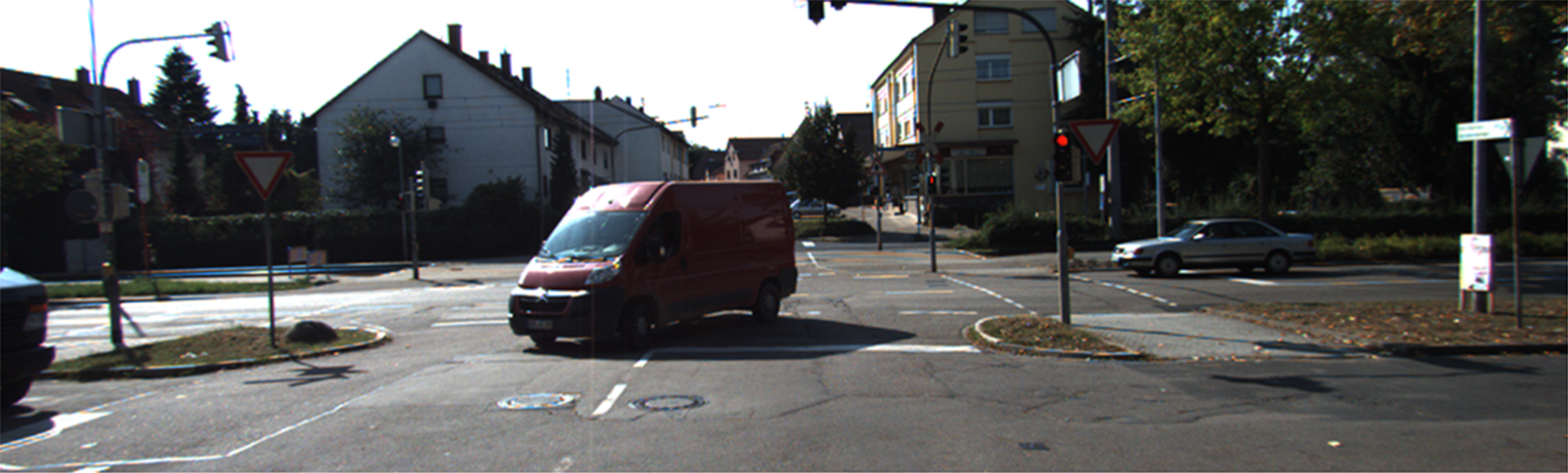}};
         \draw[red, thick] (1.75, 0.35) rectangle (5.5, 1.2);
        \end{tikzpicture} 
         &
         \begin{tikzpicture}
         \node[anchor=south west, inner sep=0] (image) at (0,0) {\includegraphics[width=0.32\linewidth]{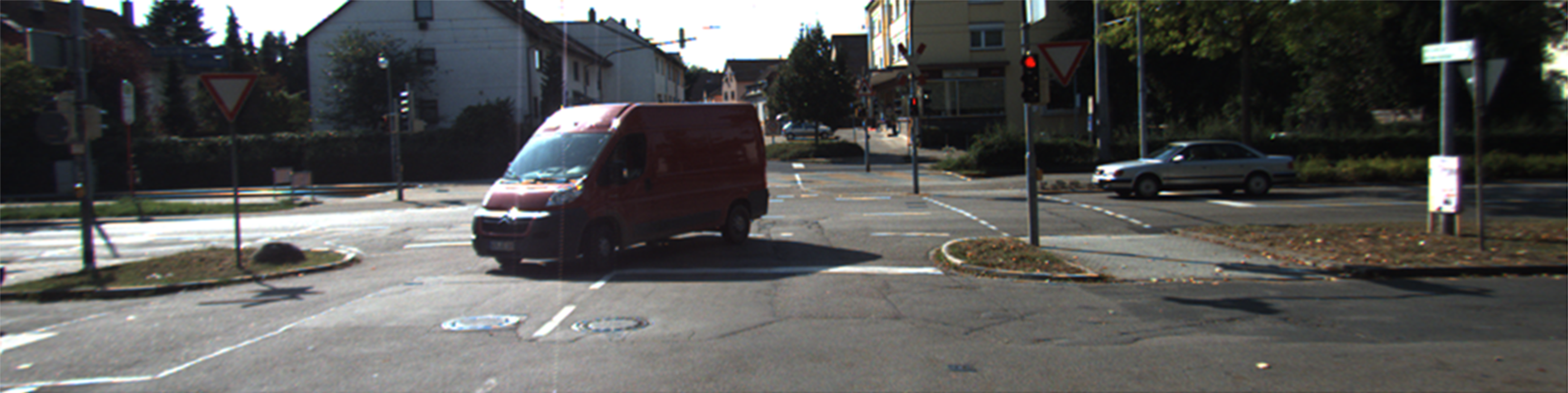}};
         \draw[red, thick] (1.75, 0.35) rectangle (5.5, 1.2);
         \end{tikzpicture} 
         &
         \includegraphics[width=0.32\linewidth]{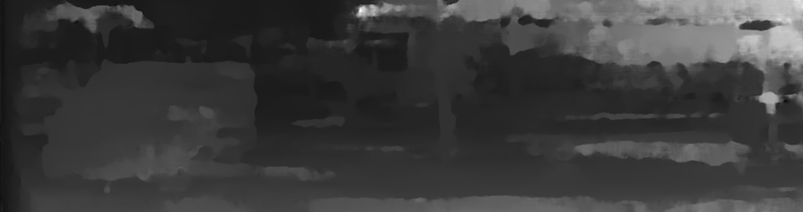}
         \\

         \multicolumn{3}{c}{After Frost Corruption Severity=1} \\
         \begin{tikzpicture}
        \node[anchor=south west, inner sep=0] (image) at (0,0) {\includegraphics[clip,trim=0cm 0cm 0cm 3.425cm, width=0.32\linewidth]{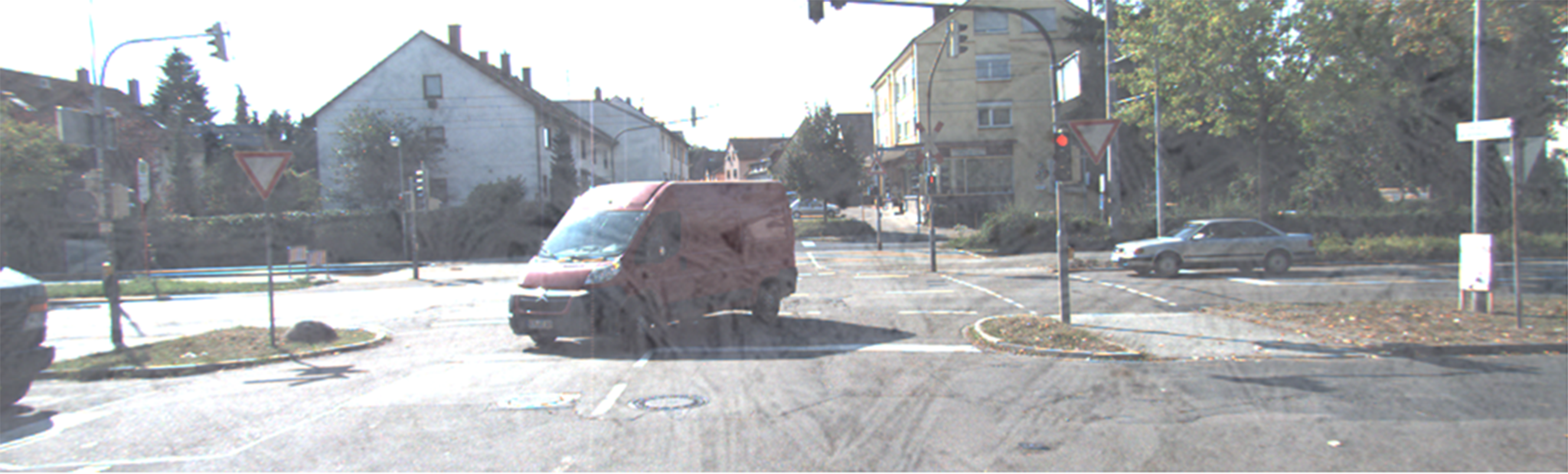}     };
        \draw[red, thick] (1.75, 0.35) rectangle (5.5, 1.2);
        \end{tikzpicture} 
         &        
         \begin{tikzpicture}
         \node[anchor=south west, inner sep=0] (image) at (0,0) {\includegraphics[width=0.32\linewidth]{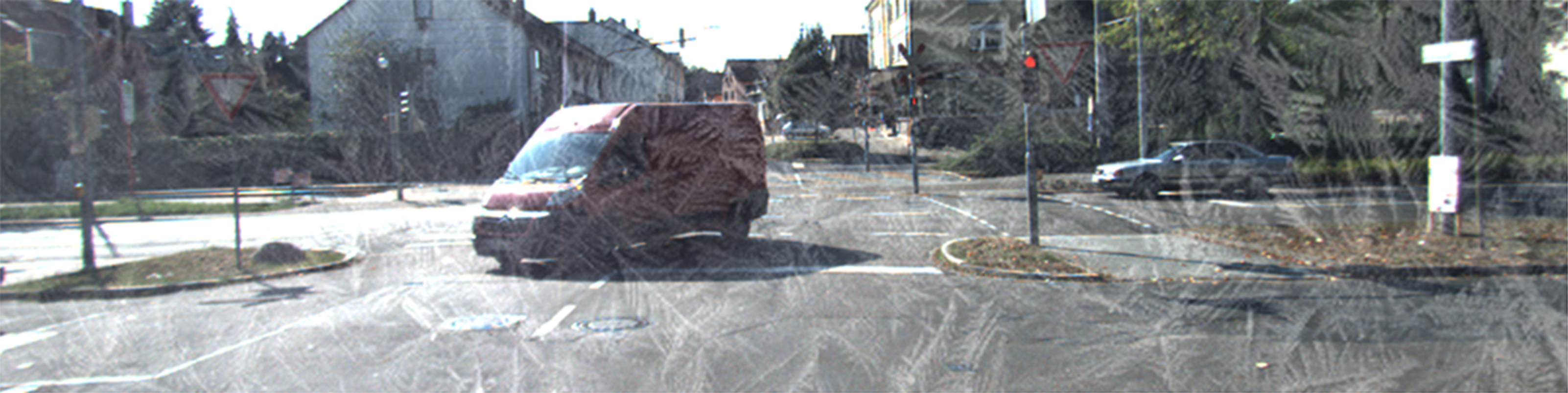}};
         \draw[red, thick] (1.75, 0.35) rectangle (5.5, 1.2);
         \end{tikzpicture} 
         &
         \includegraphics[width=0.32\linewidth]{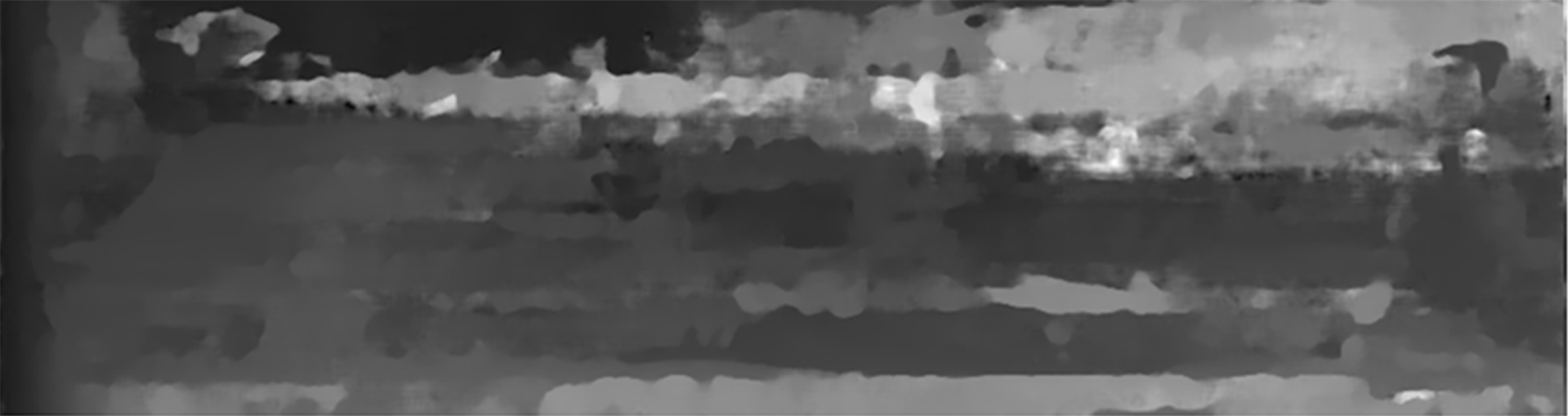}
         \\

         \multicolumn{3}{c}{After Frost Corruption Severity=3} \\
         \begin{tikzpicture}
        \node[anchor=south west, inner sep=0] (image) at (0,0) {\includegraphics[clip,trim=0cm 0cm 0cm 3.425cm, width=0.32\linewidth]{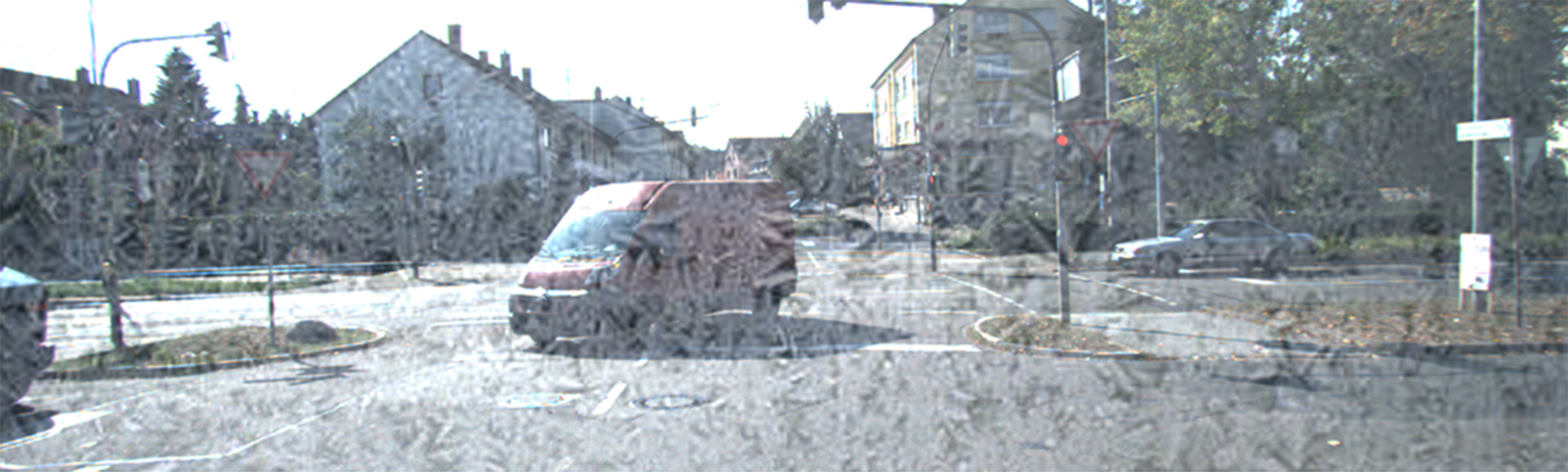}     };
        \draw[red, thick] (1.75, 0.35) rectangle (5.5, 1.2);
        \end{tikzpicture} 
         &
         \begin{tikzpicture}
         \node[anchor=south west, inner sep=0] (image) at (0,0) {\includegraphics[width=0.32\linewidth]{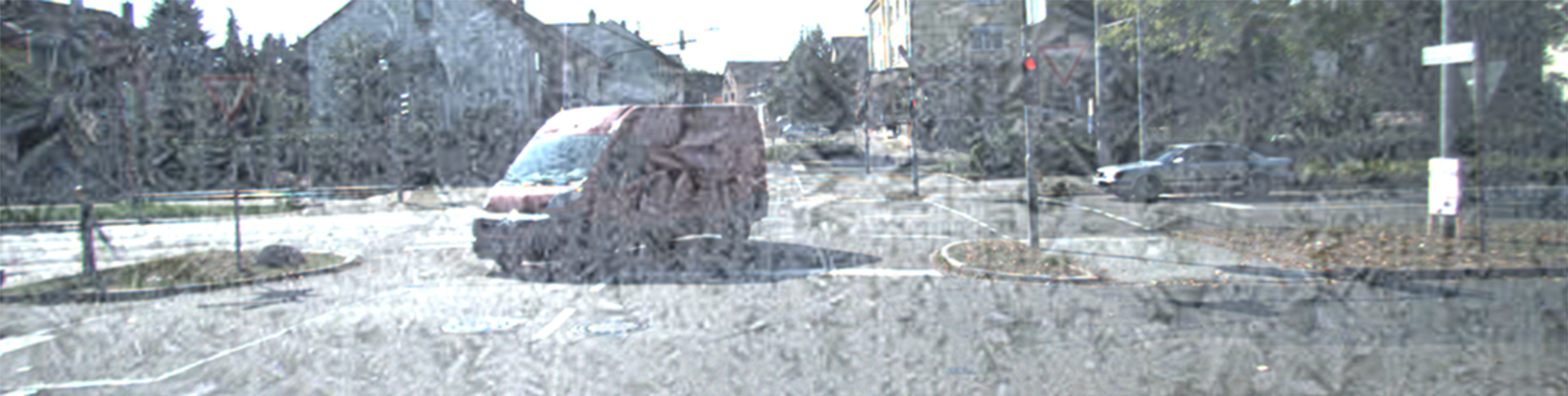}};
         \draw[red, thick] (1.75, 0.35) rectangle (5.5, 1.2);
         \end{tikzpicture} 
         &
         \includegraphics[width=0.32\linewidth]{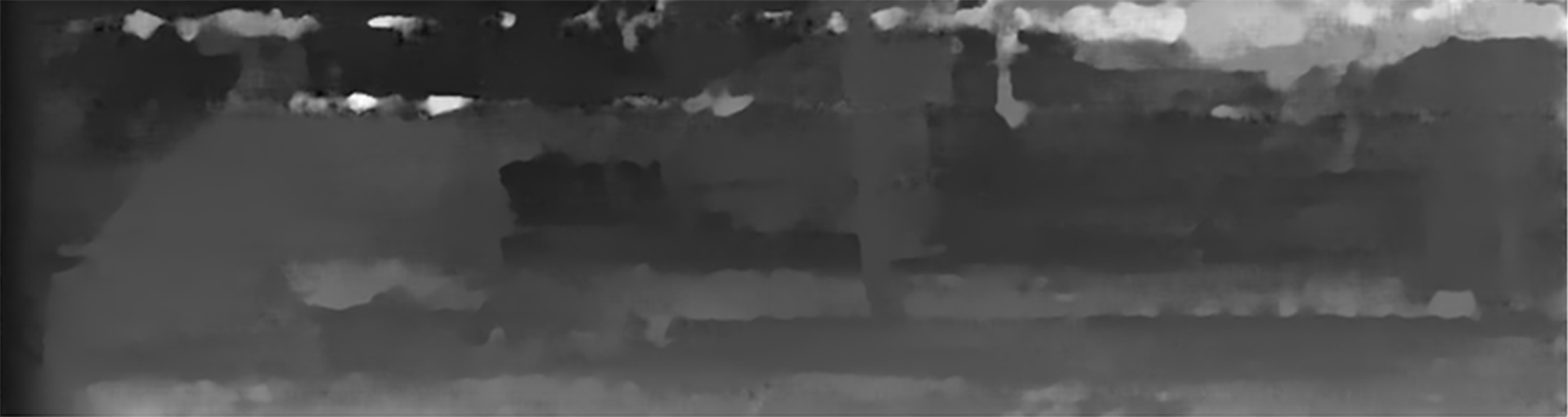}
         \\

         \multicolumn{3}{c}{After Frost Corruption Severity=5} \\
         \begin{tikzpicture}
        \node[anchor=south west, inner sep=0] (image) at (0,0) {\includegraphics[clip,trim=0cm 0cm 0cm 3.425cm, width=0.32\linewidth]{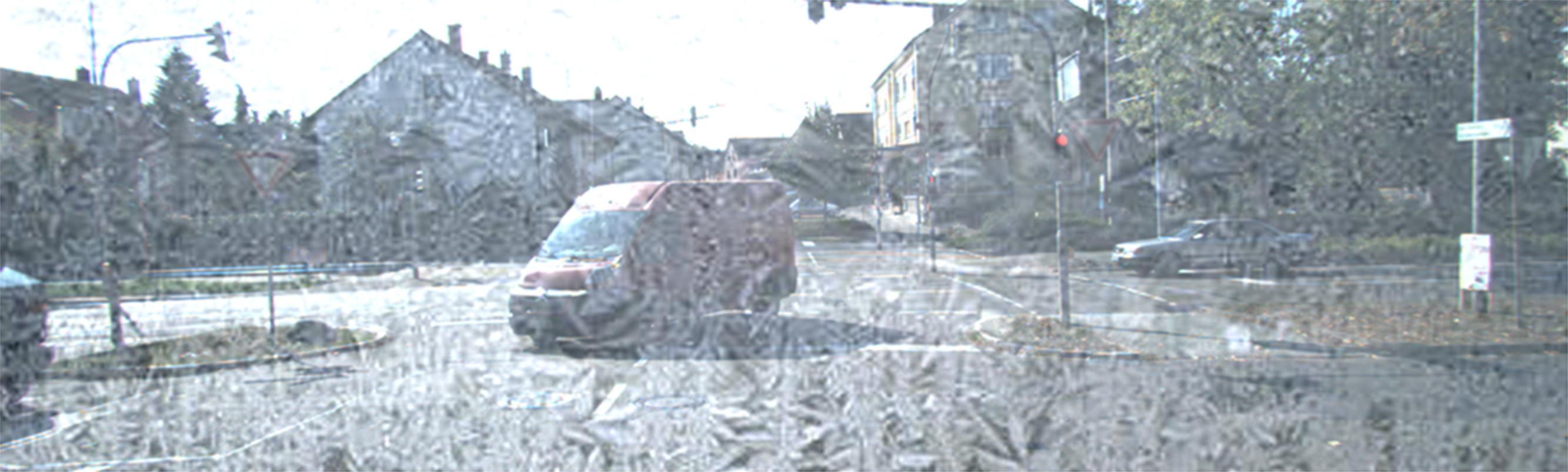}     };
        \draw[red, thick] (1.75, 0.35) rectangle (5.5, 1.2);
        \end{tikzpicture} 
         &
         \begin{tikzpicture}
         \node[anchor=south west, inner sep=0] (image) at (0,0) {\includegraphics[width=0.32\linewidth]{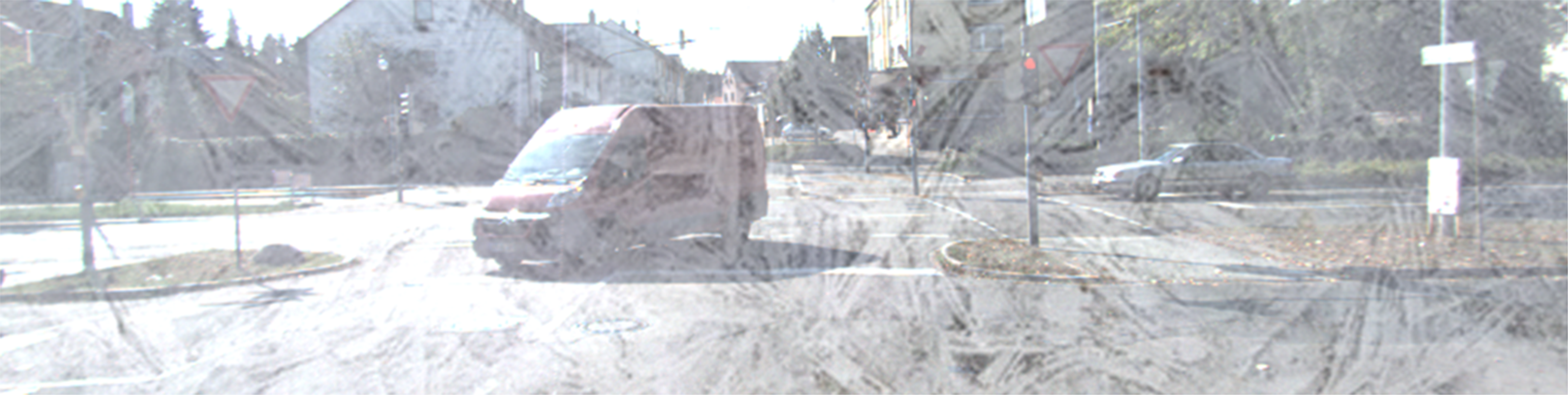}};
         \draw[red, thick] (1.75, 0.35) rectangle (5.5, 1.2);
         \end{tikzpicture} 
         &
         \includegraphics[width=0.32\linewidth]{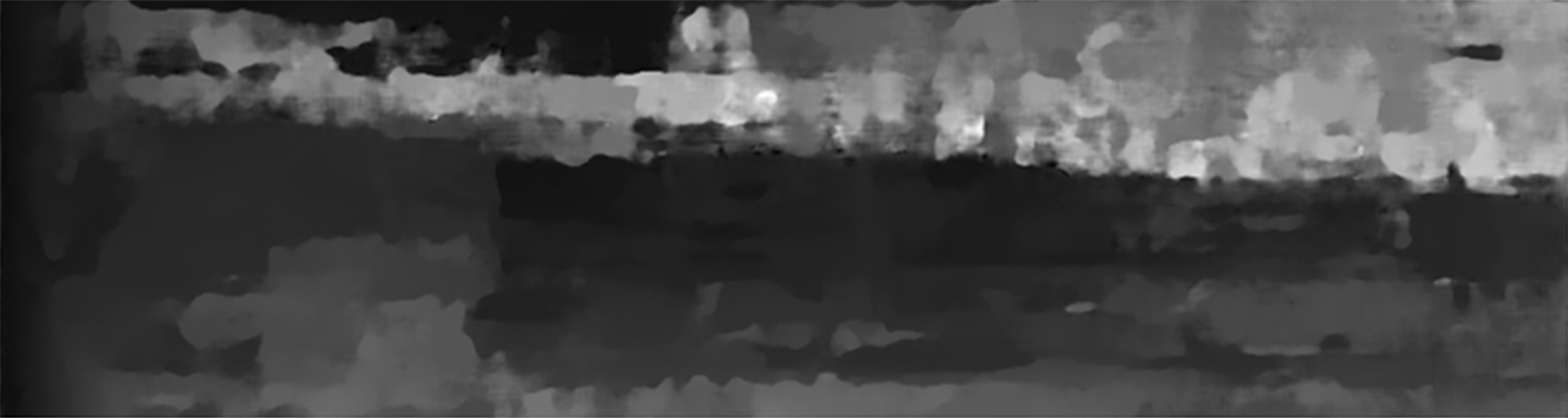}
         \\
        
    \end{tabular}
    }
    \caption{Example of predictions using STTR on KITTI2015 dataset under different severities of the 2D Common Corruption: Frost.}
    \label{fig:2dcc_exp_kitti_sttr}
\end{figure*}

The adversarial attacks supported by \benchmark{} are \textit{FGSM, BIM, PGD, APGD, and CosPGD}.

In \cref{fig:adv_exp_kitti_sttr}, we show example images perturbed using different adversarial attacks and the change in disparity estimation performed by STTR. 
Here, all attacks are optimized for 20 attack iterations, with $\alpha$=0.01 and $\epsilon=\frac{8}{255}$ under the $\ell_{\infty}$-norm bound.

\subsection{2D Common Corruptions}
\label{subsec:evaluation_details:2dcc}
To evaluate a model for a given dataset with 2D Common Corruptions, the following lines of code are required.
\begin{minted}[fontsize=\small, breaklines]{python}
from dispbench.evals import evaluate
model, results = evaluate( 
 model_name='STTR', dataset='KITTI2015', retrieve_existing=True,
 threat_config='config.yml')
\end{minted}
Here, the `config.yml' contains the configuration for the threat model; for example, when the threat model is 2D Common Corruption, `config.yml' could contain `threat\_model=\textit{``2DCommonCorruption''}', and  `severity=\textit{3}'.
Please note, when the `threat\_model' is the common corruption, \benchmark{} performs evaluations on all corruptions under the respective `threat\_model' and returns the method's performance on each corruption at the requested severity.
The argument description is as follows:
\begin{itemize}    
\item `model\_name' is the name of the disparity estimation method to be used, given as a string.
\item `dataset' is the name of the dataset to be used also given as a string. 
\item `retrieve\_existing' is a boolean flag, which when set to `True' will retrieve the evaluation from the benchmark if the queried evaluation exists in the benchmark provided by this work, else \benchmark{} will perform the evaluation.
If the `retrieve\_existing' boolean flag is set to `False' then \benchmark{} will perform the evaluation even if the queried evaluation exists in the provided benchmark.
\item The `config.yml' contains the following:
\begin{itemize}
    \item `threat\_model' is the name of the common corruption to be used, given as a string, i.e.~`2DCommonCorruption'.
    \item `severity' is the severity of the corruption, given as an integer between 1 and 5 (both inclusive).
\end{itemize}
\end{itemize}
\benchmark{} supports the following 2D Common Corruption: `gaussian\_noise', shot\_noise', `impulse\_noise', `defocus\_blur', `frosted\_glass\_blur', `motion\_blur', `zoom\_blur', `snow', `frost', `fog', `brightness', `contrast', `elastic', `pixelate', `jpeg'.
For the evaluation, \benchmark{} will evaluate the model on the validation images from the respective dataset corrupted using each of the aforementioned corruptions for the given severity and then report the mean performance over all of them.

In \cref{fig:2dcc_exp_kitti_sttr}, we show example images perturbed using the 2D Common Corruption: Frost and the change in disparity estimation performed by STTR over different severity strengths.

\subsection{Dataset Details}
\label{sec:appendix:dataset_details}
\benchmark{} currently supports two distinct disparity datasets.
Following, we describe these datasets in detail.

\subsubsection{FlyingThings3D}
\label{subsec:appendix:dataset_details:flyingthings3d}
This is a synthetic dataset proposed by \cite{flyingthings_dispnet} largely used for training and evaluation of disparity estimation methods.
This dataset consists of 25000 stereo frames, of everyday objects such as chairs, tables, cars, etc. flying around in 3D trajectories.
The idea behind this dataset is to have a large volume of trajectories and random movements rather than focus on a real-world application.
In their work, \cite{flownet} showed models trained on FlyingThings3D can generalize to a certain extent to other datasets.

\subsubsection{KITTI2015}
\label{subsec:appendix:dataset_details:kitti2015}
Proposed by \cite{kitti15}, this dataset is focused on the real-world driving scenario.
It contains a total of 400 pairs of image frames, split equally for training and testing.
The image frames were captured in the wild while driving around on the streets of various cities.
The ground-truth labels were obtained by an automated process.

%%%%%%%%%%%%%%%%%%%%%%%%%%%%%    END METHODS    %%%%%%%%%%%%%%%%%%%%%

%_________________________    START METRICS    ____________________

\section{Initial Evaluations using \benchmark{}}
\label{sec:analysis} 
\begin{figure*}[h]
    \centering
    \includegraphics[width=1.0\linewidth]{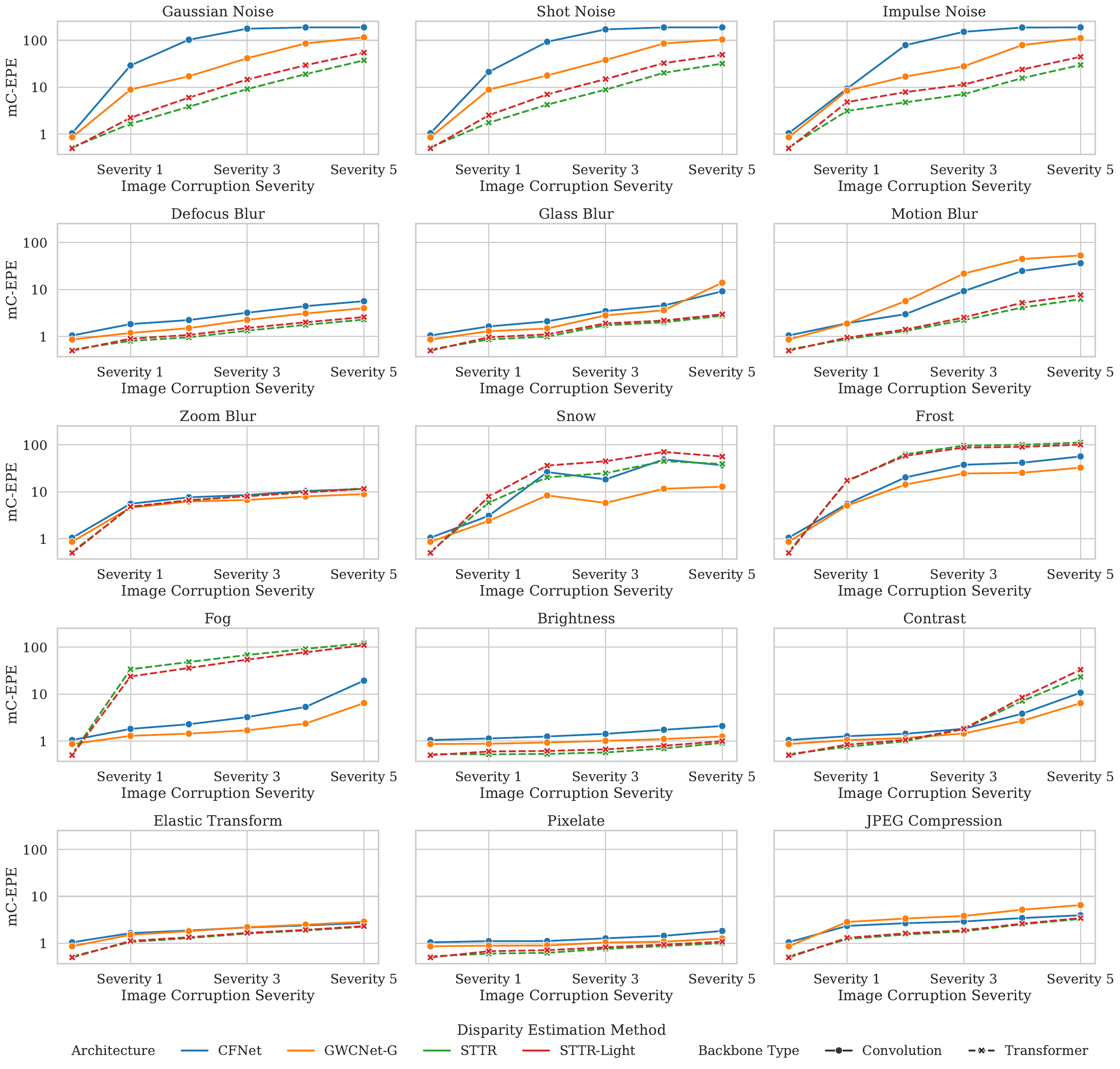}
    \caption{Using the FlyingThings3D dataset for disparity estimation, we perform an initial benchmarking of i.i.d. performance and generalization abilities of four popular disparity estimation methods. CFNet and GWCNet are traditional CNN-based stereo matching methods, whereas STTR and STTR-light are newly proposed transformer-based large models capable of zero-shot disparity estimation. Here, we use their fine-tuned versions for the FlyingThings3D dataset. The y-axis reports the mean EPE over the entire validation set for the respective corruption, and the x-axis denotes the severity of the 2D Common Corruption used to corrupt the input images. We report the i.i.d. performance at severity=0. Here we observe that while all four methods are highly vulnerable to Noise and Weather corruptions, newly proposed STTR and STTR-light are surprisingly less robust than the older CNN-based methods against weather corruptions. This finding is interesting and concerning as weather corruptions are the most likely real-world domain shift.}
    \label{fig:2dcc_perf_sceneflow}
\end{figure*}
%\section{Generalization Ability Measure}
%\label{sec:metrics:generalization}
We use \benchmark{} to perform some initial benchmarking and make some interesting observations.
Following, we discuss the details of the benchmarking process.
Please note, we use the FlyingThing3D and the KITTI2015 dataset for the benchmarking.
However, very few pretrained architectures are available for KITTI2015, and thus our evaluations using KITTI2015 are limited to these.
While \benchmark{} enables the training of architectures of multiple datasets, doing so is beyond our resource capabilities.

For additional details on the datasets, please refer to \cref{sec:appendix:dataset_details}.

\noindent\paragraph{Measuring Generalization Ability. }
Inspired by multiple works~\citep{robustbench,hendrycks2020augmix,hoffmann2021towards} that use OOD Robustness of methods for evaluating the generalization ability of the method, even evaluate over every common corruption, that is the 15 2D Common Corruptions: `Gaussian Noise', Shot Noise', `Impulse Noise', `Defocus Blur', `Frosted Glass Blur', `Motion Blur', `Zoom Blur', `Snow', `Frost', `Fog', `Brightness', `Contrast', `Elastic Transform', `Pixelate', `JPEG Compression'.
Then, we find the mean EPE w.r.t. the ground truth for a given method, across all corruptions at a given severity and report use this to measure the Generalization Ability.
We corrupt the pair of stereo images with the same corruption at the same severity when evaluating. 

Ideally, one would like to evaluate the generalization ability and reliability of methods using real-world samples captured in the wild.
However, annotation of these samples is a challenging and time-consuming task, and thus, no such dataset is available for disparity estimation.
\citet{acdc} captured such data in the wild with domain shifts due to changes in time of day and changes in weather conditions like snowfall, rain, and fog.
They also provide pixel-level annotations for their images, however, these annotations are only available for semantic segmentation, and these images are monocular and not stereo.
They propose this as the  Adverse Conditions Dataset with Correspondences for Semantic Driving Scene Understanding (ACDC) dataset.
Interestingly, in their work, \citet{anonymous_semseg} showed a very strong positive correlation between the performance of most methods on the ACDC dataset and their performance against in-domain images corrupted with the 2D Common Corruptions to cause a synthetic domain shift.
This is an important finding as it proves that 2D Common Corruptions can be used as a proxy to real-world domain shifts.
We discuss this in \cref{sec:appendix:semseg_acdc_cc}.
For details on the dataset, please refer to the appendix.

\noindent\paragraph{Measuring Reliability Under Adversarial Attacks. }
Adversarial attacks, especially white-box attacks, serve as a proxy to the worst-case scenario and help understand the quality of the representations learned by a model~\cite{agnihotri2023cospgd,schmalfuss2022perturbationconstrained,apgd}.
\benchmark{} provides the ability to evaluate the models against some popular adversarial attacks, as discussed in \cref{subsec:evaluation_details:adv_attack}.
However, we focus this work towards realistic corruptions possible in the real world.
For evaluations over adversarial attacks, please refer to \cref{sec:appendix:analyis_extension}.

%%%%%%%%%%%%%%%%%%%%%%%%%%%%%    END METRICS    %%%%%%%%%%%%%%%%%%%%%
\noindent\paragraph{Architectures Used. }
%We use STTR, STTR-light, .....
Disparity estimation networks essentially estimate optimal correspondence matching between pixels on epipolar lines in the left and right images to infer depth. 
Most disparity estimation architectures used a cost volume with cross-correlation or contamination of feature representations for the left and right images.
However, \textbf{GWCNet-G}~\cite{gwcnet} proposed using group-wise correlations to construct the cost volume.
This leads to a significant boost in i.i.d.~performance and inference speed.
\textbf{CFNet}~\cite{cfnet} proposed fusing on multiple low-resolution dense cost volumes to enlarge the receptive field, enabling extraction of robust structural representations, followed by cascading the cost volume representations to alleviate the unbalanced disparity estimation. It was proposed to be robust to large domain differences and was SotA when proposed.
\textbf{Stereo-Transformers (STTR)}, \citet{sttr} proposes to replace the cost volume construction with dense pixel matching using position information and attention to enable sequence-to-sequence matching. This relaxes the limitation of a fixed disparity range and identifies occluded regions with confidence estimates. 
STTR generalizes across different domains, even without fine-tuning.
However, in our evaluations, we use fine-tuned checkpoints for a fair comparison of reliability and generalization capabilities. 
\textbf{STTR-light} is the lightweight version of STTR proposed for faster inference with only a marginal drop in i.i.d.~performance.
We use the publicly available pre-trained checkpoints for our evaluations.
%_________________________    START ANALYSIS    ____________________
\section{Key Findings}
\label{subsec:analysis:findings}
In the following, we present the key findings made using the initial benchmarking using the \benchmark{}.

\subsection{FlyingThings3D}
\label{subsec:analysis:findings:flything3d}
Following, we discuss the observations made in the robustness benchmark created using \benchmark{}.
We report the evaluations in \cref{fig:2dcc_perf_sceneflow}.
Here, we observe that indeed the i.i.d. performance of the new methods like STTR and STTR-light is better than the older CNN-based CFNet and GWCNet-G, however, the same is not always true for their generalization abilities.
All four considered methods appear to be robust to digital corruptions such as changes in brightness, contrast, elastic transform, pixelated, and JPEG compression to a significant extent. 
While, all four methods appear to be extremely non-robust to additive noise, possible in the real world due to sensor error, causing the mean errors to go as high as 100. 
Please note that compared to the single-digit EPE values for i.i.d. performance, these errors are significantly high.

The most interesting behavior is seen under different weather corruptions: Snow, Frost, and Fog.
Here, all four methods are non-robust, however, the newer transformer-based methods STTR and STTR-light are significantly more non-robust.
This is quite alarming, as weather corruptions are the most natural domain shifts possible in the real world, and here, the large models fail significantly worse. 
Especially under Frost and Fog corruptions, the larger STTR performs worse than its lightweight counterpart, STTR-light.
This raises some interesting concerns that warrant further study and deeper analysis.

\subsection{KITTI2015}
\label{subsec:analysis:findings:kitti}
\begin{figure*}[h]
    \centering
    \includegraphics[width=1.0\linewidth]{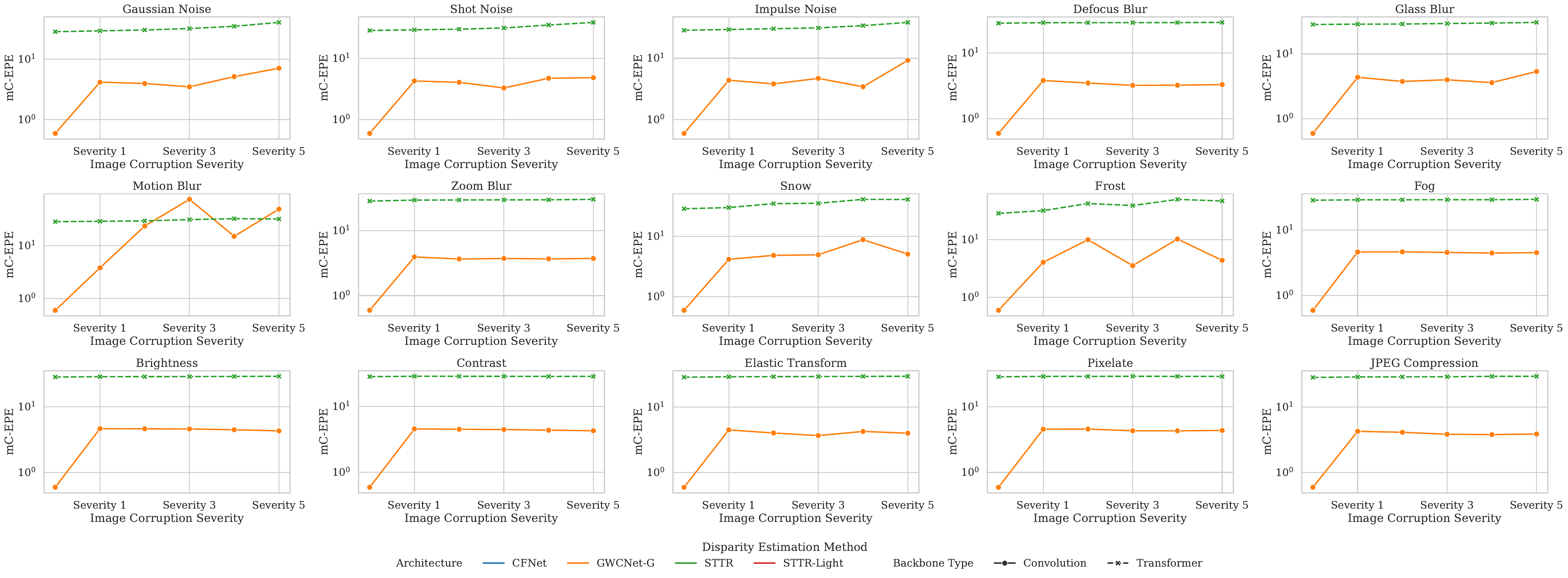}
    \caption{Using the KITTI2015 dataset for disparity estimation, we perform an initial benchmarking of i.i.d. performance and generalization abilities of the two popular and available disparity estimation methods. GWCNet is a traditional CNN-based stereo matching method, whereas STTR is a newly proposed transformer-based large model capable of zero-shot disparity estimation. Here, we use their fine-tuned versions for the KITTI2015 dataset. The y-axis reports the mean EPE over the entire validation set for the respective corruption, and the x-axis denotes the severity of the 2D Common Corruption used to corrupt the input images. We report the i.i.d. performance at severity=0. Here, we observe that while both the methods are highly vulnerable to Noise and Weather corruptions, the newly proposed STTR is surprisingly less robust than the older CNN-based method against all corruptions. This finding is interesting and concerning as it contradicts the findings on the Synthetic Dataset FlyingThings3D in \cref{fig:2dcc_perf_sceneflow}.}
    \label{fig:2dcc_perf_kitti}
\end{figure*}
There are very limited pre-trained architectures available on KITTI2015 for the disparity estimation task, namely GWCNet-G and STTR.
We perform our analysis using these and report the evaluations in \cref{fig:2dcc_perf_kitti}.
We observe that the newly proposed STTR is less robust than GWCNet-C across all corruptions and severities.
This does not align with the observations made with synthetic corruptions on the synthetic dataset FlyingThings3D.
This suggests that further analysis is required.

\section{Synthetic Corruptions on Synthetic Dataset vs Synthetic Corruptions on Real World Dataset}
\label{subsec:analysis:findings:synthetic_vs_real}
\begin{figure*}[h]
    \centering
    \includegraphics[width=1.0\linewidth]{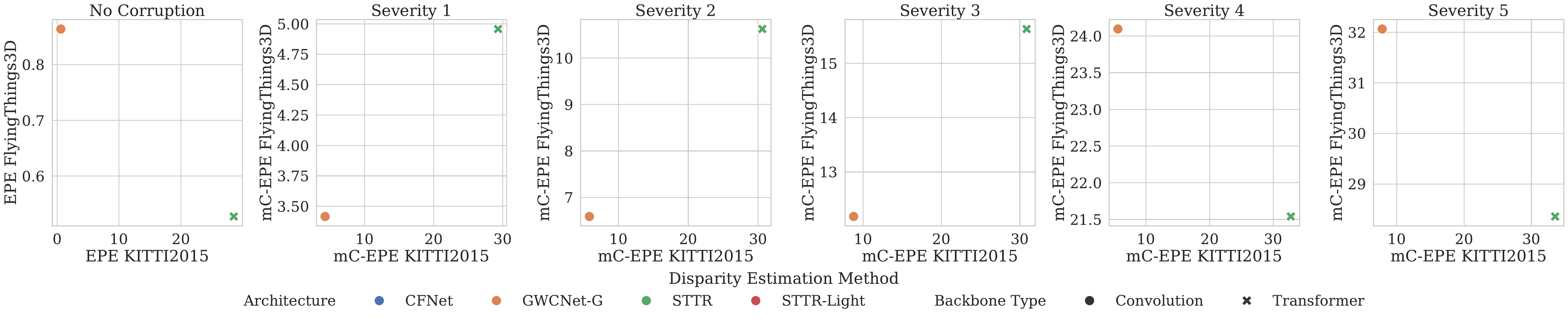}
    \caption{For the same architecture, we evaluate checkpoint pretrained on Flyingthings3D against synthetic 2D Common Corruption on Flyingthings3D and correlate its performance with the checkpoint trained on KITTI2015 against synthetic 2D Common Corruptions on KITTI2015.
    Here, we report the mean EPE across all corruptions for a given severity level. 
    For individual corruptions, please refer to \cref{fig:all_correlation_synthetic_real}.
    We observe no correlation in performance, indicating that synthetic corruptions on synthetic datasets cannot be used as a proxy for real-world corruptions.}
    \label{fig:mean_correlation_synthetic_real}
\end{figure*}
Following the findings from \cref{subsec:analysis:findings:kitti}, we investigate whether the performance of models on synthetic corruptions (2D Common Corruptions) on synthetic dataset (FlyingThings3D) can serve as a proxy to the performance of models on synthetic corruptions (2D Common Corruptions) on real-world data (KITTI2015).
We report this analysis in \cref{fig:mean_correlation_synthetic_real} and observe that 
synthetic corruptions on synthetic datasets do not represent synthetic corruptions on real-world datasets.
As known from \cite{anonymous_semseg}, synthetic corruptions on real-world datasets represent real-world corruptions.
By extension, synthetic corruptions on synthetic datasets do not represent real-world corruptions.
This crucial finding eliminates the possibility of using synthetic simulators like CARLA~\cite{carla}, LGSVL (SVL Simulator)~\cite{rong2020lgsvl}, AirSim~\cite{airsim2017fsr}, and others for possible applications in the real world. 
%%%%%%%%%%%%%%%%%%%%%%%%%%%%%    END ANALYSIS    %%%%%%%%%%%%%%%%%%%%%
%_________________________    START DISCUSSION    ____________________

\section{Conclusion}
\label{sec:conclusion}
Evaluating a model's robustness is vital for real-world applications.
However, capturing corruptions in the real world is time and resource intensive.
Here, synthetic corruptions appear to be an attractive alternative.
Thus, we propose \benchmark{}, the first robustness benchmarking tool and a novel benchmark on synthetic corruptions for disparity estimation methods.
We discuss the unique features of \benchmark{} in detail and demonstrate that the library is user-friendly, such that adding new methods or performing evaluation is very intuitive.
We use \benchmark{} to evaluate the i.i.d. performance and OOD generalization of some popularly used disparity estimation methods.
We observe that under realistic scenarios, recently proposed large transformer-based methods known to be SotA on i.i.d. samples do not generalize well to image corruptions, demonstrating the gap in current research when considering real-world applications.
Lastly, we show experimentally that synthetic corruptions on synthetic datasets do not represent real-world corruptions, thus, synthetic corruptions on real-world datasets present a more promising path.
\benchmark{} enables a more in-depth understanding of the reliability and generalization abilities of disparity estimation methods, and its consolidated nature would make future research more streamlined.

%\subsection{Future Work}
%\label{subsec:conclusion:future}
\noindent\paragraph{Future Work. }
Very recently, OpenStereo~\cite{OpenStereo} has been made public that supports newly proposed stereo matching methods, which are foundational models for stereo matching like StereoAnything~\cite{guo2024stereoanything}, and LightStereo~\cite{guo2024lightstereo}.
We intend to adapt our evaluator into OpenStereo to enable safety studies of SotA disparity estimation methods.
Additionally, more in-depth analysis of the disparity estimation methods, for example, as done by \cite{gavrikov2024training} for classification methods, would help understand the models and their workings, especially in terms of their robustness performance.

%\subsection{Limitations}
%\label{subsec:conclusion:limitations}
\noindent\paragraph{Limitations. }
Benchmarking disparity estimation methods is a compute and labor-intensive endeavor. 
Thus, best utilizing available resources, we currently use \benchmark{} to benchmark a limited number of settings, using the most popular works for now.
The benchmarking tool itself offers significantly more combinations that can be benchmarked.
%Very recently OpenStereo~\cite{OpenStereo} has been made public that supports newly proposed stereo matching methods foundational models for stereo matching like StereoAnything~\cite{guo2024stereoanything}, and its lightweight version LightStereo~\cite{guo2024lightstereo}.
%We intend to adapt our evaluator into OpenStereo to enable reliability and safety studies of disparity estimation methods with SotA methods.

\section*{Reproducibility Statement}
Every experiment in this work is reproducible and is part of an effort toward open-source work.
\benchmark{} will be open-source and publicly available, including all evaluation logs and model checkpoint weights. 
This work intends to help the research community use synthetic corruptions to build more reliable and generalizable disparity estimation methods such that they are ready for deployment in the real world even under safety-critical applications.
The open-source code and model weights for \benchmark{} is available here: \url{https://github.com/shashankskagnihotri/benchmarking_robustness/tree/disparity_estimation/final/disparity_estimation}.

\section*{Acknowledgments}
M.K. acknowledges funding by the DFG Research Unit 5336. The authors acknowledge support by the state of Baden-Württemberg through bwHPC.

\iffalse
\subsubsection*{Author Contributions}
If you'd like to, you may include a section for author contributions, as is done
in many journals. This is optional and at the discretion of the authors.

\fi

{
    \small
    \bibliographystyle{ieeenat_fullname}
    \bibliography{main}
}
\newpage
\appendix
\onecolumn
{
    \centering
    \Large
    \textbf{\benchmark{}: Benchmarking Disparity Estimation to Synthetic Corruptions} \\
    \vspace{0.5em}Paper \#6 Supplementary Material \\
    %\vspace{0.5em} Supplementary Material \\
    \vspace{1.0em}
}

\section*{Table Of Content}
The supplementary material covers the following information:
\begin{itemize}
\setlength\itemsep{2em}
    \item \Cref{sec:appendix:semseg_acdc_cc}: We show that synthetic 2D Common Corruptions indeed serve as a proxy to domain shifts in the real world.
    %\item \Cref{sec:appendix:dataset_details}: Details for the datasets used.
    %\begin{itemize}
    %    \item %\Cref{subsec:appendix:dataset_details:flyingthings3d}: %FlyingThings3D
    %    \item %\Cref{subsec:appendix:dataset_details:kitti2015}: %KITTI2015
    %\end{itemize}
    \item \Cref{sec:appendix:implementation_details}: Additional implementation details for the evaluated benchmark.
    \item \Cref{sec:appendix:description}: In detail description of the attacks.
    \item \Cref{sec:appendix:model_zoo}: \benchmark{} function call to get model weights.
    \item \Cref{sec:appendix:evaluation_details}: In detail explanation of the available functionalities of the \benchmark{} benchmarking tool and description of the arguments for each function.
    \item \Cref{sec:appendix:analyis_extension}: Here we provide additional results from the benchmark evaluated using \benchmark{}. 

    %\item \textcolor{red}{At the end, we attach the anonymous paper of \cite{anonymous_semseg} for the ease of the reviewer.}
\end{itemize}
\vspace{1em}
Please note that due to the similarity of the objective, many aspects of this appendix are very similar to \citet{anonymous_semseg}.

\section{Do Synthetic Corruptions Represent The Real World?}
\label{sec:appendix:semseg_acdc_cc}
\begin{figure}
    \centering
    \includegraphics[width=1.0\linewidth]{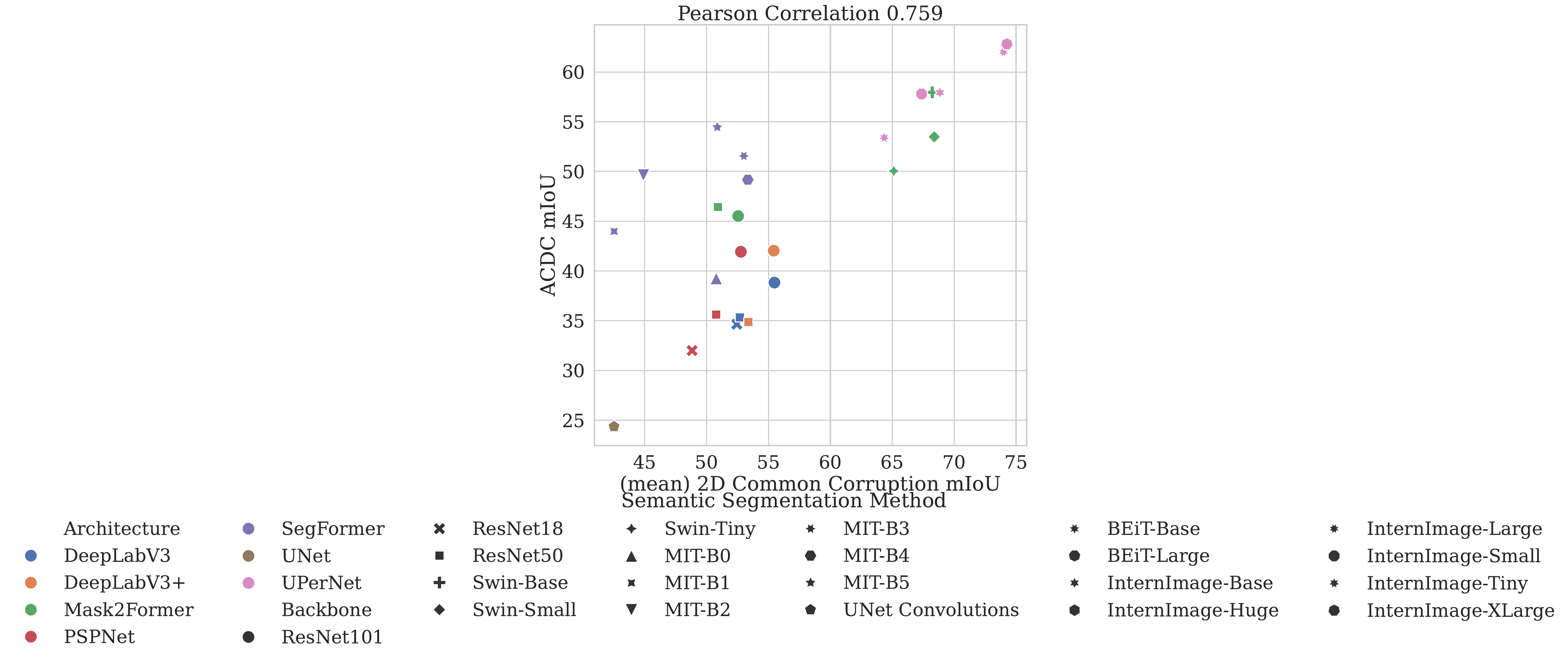}
    \caption{\textcolor{black}{Results from work by \citet{anonymous_semseg}. Here they find a \textbf{very strong positive correlation between mean mIoU over the ACDC evaluation dataset~\citep{acdc} and mean mIoU over each 2D Common Corruption~\citep{commoncorruptions}} over the Cityscapes dataset~\citep{cordts2016cityscapes}. All models were trained using the training subset of the Cityscapes dataset. ACDC is the Adverse Conditions Dataset with Correspondences for Semantic Driving Scene Understanding captured in similar scenes are cityscapes but under four different domains: Day/Night, Rain, Snow, and Fog in the wild. ACDC is a community-used baseline for evaluating the performance of semantic segmentation methods on domain shifts observed in the wild.}}
    \label{fig:semseg_acdc_cc}
\end{figure}
\textcolor{black}{In their work \citet{anonymous_semseg}, they find the correlation between mean mIoU over the ACDC evaluation dataset~\citep{acdc} and mean mIoU over each 2D Common Corruption~\citep{commoncorruptions} over the Cityscapes dataset~\citep{cordts2016cityscapes}. 
We include \Cref{fig:semseg_acdc_cc} from their work here for ease of understanding.
All models were trained using the training subset of the Cityscapes dataset. 
ACDC is the Adverse Conditions Dataset with Correspondences for Semantic Driving Scene Understanding captured in similar scenes are cityscapes but under four different domains: Day/Night, Rain, Snow, and Fog in the wild. ACDC is a community-used baseline for evaluating the performance of semantic segmentation methods on domain shifts observed in the wild.
They find that there exists a very strong positive correlation between the two.
This shows, that \textbf{yes, synthetic corruptions can serve as a proxy for the real world}.
Unfortunately, a similar ``in the wild'' captured dataset does not exist for optical flow estimation to evaluate the effect of domain shifts on the performance of optical flow methods. 
However, given that for the task of semantic segmentation, we find a very high positive correlation between the performance on real-world corruptions and synthetic corruptions, it is a safe assumption that the same would hold true for optical flow estimation as well.
Thus, in this work, we evaluate against synthetic 2D Common Corruptions~\citep{commoncorruptions} and synthetic 3D Common Corruptions~\citep{3dcommoncorruptions}.
}

\section{Implementation Details Of The Benchmark}
\label{sec:appendix:implementation_details}
Following, we provide details regarding the experiments done for creating the benchmark used in the analysis.

\noindent\paragraph{Compute Resources. }Most experiments were done on a single 40 GB NVIDIA Tesla V100 GPU each, however, MS-RAFT+, FlowFormer, and FlowFormer++ are more compute-intensive, and thus 80GB NVIDIA A100 GPUs or NVIDIA H100 were used for these models, a single GPU for each experiment.

\noindent\paragraph{Datasets Used. }Performing adversarial attacks and OOD robustness evaluations are very expensive and compute-intensive.
Thus, performing evaluation using all model-dataset pairs is not possible given the limited computing resources at our disposal.
Thus, for the benchmark, we only use FlyingThings3D and KITTI2015, as these are the most commonly used datasets for evaluation~\citep{guo2024stereoanything,sttr,cfnet,flyingthings_dispnet}.

\noindent\paragraph{Metrics Calculation. } In this work, for robustness evaluations we consider the mC-EPE, which is the mean End-Point-Error of a method, against common corruptions at a given severity, over every input image from the validation dataset.
We use all 15 2D Common Corruptions: `Gaussian Noise', Shot Noise', `Impulse Noise', `Defocus Blur', `Frosted Glass Blur', `Motion Blur', `Zoom Blur', `Snow', `Frost', `Fog', `Brightness', `Contrast', `Elastic Transform', `Pixelate', `JPEG Compression'.
All the common corruptions are at severity=$\{1,\text{ }2,\text{ }3,\text{ }4,\text{ }5\}$.
\cite{3dcommoncorruptions} offers more 3D Common Corruptions, however computing them is resource intensive. 
Thus, given our limited resources and an overlap in the corruptions between 2D Common Corruptions and 3D Common Corruptions, we focus on generating 3D Common Corruptions for now, however, we intend to extend \benchmark{} to also evaluate on the 3D Common Corruptions.

\textbf{Calculating the EPE. } $EPE$ is the Euclidean distance between the two vectors, where one vector is the predicted flow by the disparity estimation method and the other vector is the ground truth in case of i.i.d. performance evaluations, non-targeted attacks evaluations, and OOD robustness evaluations, while it is the target flow vector, in case of targeted attacks.
For each dataset, the $EPE$ value is calculated over all the samples of the evaluation set of the respective dataset and then the mean $EPE$ value is used as the mean-$EPE$ of the respective method over the respective dataset.

%\textbf{Other Metrics. } Apart from EPE, \benchmark{} also enables calculating a lot of other interesting metrics, such as $\ell_1$, $\ell_2$, and cosine distance between two vectors. These vectors are the same as that in the case of $EPE$ calculations.

\section{Description of \benchmark{}}
\label{sec:appendix:description}
Following, we describe the benchmarking tool, \benchmark{}.
There exists no standardized tool for evaluating the performance of disparity estimation methods.
Thus, the codebase for such a tool had to be written from scratch.
In the following, we describe the benchmarking tool, \benchmark{}.
Currently it supports 4 unique architectures (new architectures to be added to \benchmark{} with time) and 3 distinct datasets, namely FlyingThings3D~\citep{flyingthings_dispnet}, KITTI2015~\citep{kitti15}, MPI Sintel~\citep{sintel1} (clean and final) (please refer \cref{sec:appendix:dataset_details} for additional details on the datasets).
It enables training and evaluations on all aforementioned datasets including evaluations using SotA adversarial attacks such as CosPGD~\citep{agnihotri2023cospgd}, and other commonly used adversarial attacks like BIM~\citep{bim}, PGD~\citep{pgd}, FGSM~\citep{fgsm}, under various Lipshitz ($l_p$) norm bounds. 
Additionally, it enables evaluations for Out-of-Distribution (OOD) robustness by corrupting the inference samples using 2D Common Corruptions~\citep{commoncorruptions}.

Following we show the basic commands to use \benchmark{}. 
We describe each attack and common corruption supported by \benchmark{} in detail in \cref{sec:appendix:description}.
It enables training and evaluations on all aforementioned datasets including evaluations using SotA adversarial attacks such as CosPGD~\citep{agnihotri2023cospgd}, and other commonly used adversarial attacks like BIM~\citep{bim}, PGD~\citep{pgd}, FGSM~\citep{fgsm}, under various lipshitz ($l_p$) norm bounds. 
Additionally, it enables evaluations for Out-of-Distribution (OOD) robustness by corrupting the inference samples using 2D Common Corruptions~\citep{commoncorruptions}.

We follow the nomenclature set by RobustBench~\citep{robustbench} and use ``threat\_model'' to define the kind of evaluation to be performed.
When ``threat\_model'' is defined to be ``None'', the evaluation is performed on unperturbed and unaltered images, if the ``threat\_model'' is defined to be an adversarial attack, for example ``PGD'', ``CosPGD'' or ``PCFA'', then \benchmark{} performs an adversarial attack using the user-defined parameters.
Whereas, if ``threat\_model'' is defined to be ``2DCommonCorruptions'' or ``3DCommonCorruptions'', the \benchmark{} performs evaluations after perturbing the images with 2D Common Corruptions and 3D Common Corruptions respectively.

If the queried evaluation already exists in the benchmark provided by this work, then \benchmark{} simply retrieves the evaluations, thus saving computation.

\subsection{Adversarial Attacks}
\label{subsec:appendix:description:adv_attacks}
Please note that due to the similarity of the objective, many aspects of this appendix are very similar to \citet{flowbench}.
\benchmark{} enables the use of many white-box adversarial attacks to help users better study the reliability of their disparity methods.
We choose to specifically include these white-box adversarial attacks as they either serve as the common benchmark for adversarial attacks in classification literature (FGSM, BIM, PGD, APGD) or they are unique attacks proposed specifically for pixel-wise prediction tasks (CosPGD).
These attacks can either be \emph{Non-targeted} which are designed to simply fool the model into making incorrect predictions, irrespective of what the model eventually predicts, or can be \emph{Targeted}, where the model is fooled to make a certain prediction.
Most attacks can be, designed to be either Targeted or Non-targeted, these include, FGSM, BIM, PGD, APGD, CosPGD, and Adversarial Weather.
In our current implementation, we are limited to Non-targeted attacks.
Following, we discuss these attacks in detail and highlight their key differences.

\noindent\paragraph{FGSM. }Assuming a non-targeted attack, given a model $f_{\theta}$ and an unperturbed input sample $\boldsymbol{X}^\mathrm{clean}$ and ground truth label $\boldsymbol{Y}$, FGSM attack adds noise $\delta$ to $\boldsymbol{X}^\mathrm{clean}$ as follows,

\begin{equation}
%\small
    \label{eqn:fgsm_attack_1}
    \boldsymbol{X}^{\mathrm{adv}} = \boldsymbol{X}^{\mathrm{clean}}+\alpha \cdot \mathrm{sign}\nabla_{\boldsymbol{X}^{\mathrm{clean}}}L(f_{\theta}(\boldsymbol{X}^{\mathrm{clean}}), \boldsymbol{Y}),
\end{equation}

\begin{equation}
    \label{eqn:fgsm_attack_2}  
    \delta = \phi^{\epsilon}(\boldsymbol{X}^{\mathrm{adv}} - \boldsymbol{X}^{\mathrm{clean}}), 
\end{equation}

\begin{equation}
\label{eqn:fgsm_attack_3}
    \boldsymbol{X}^{\mathrm{adv}} = \phi^{r}(\boldsymbol{X}^{\mathrm{clean}}+ \delta).
\end{equation}
Here,  $L(\cdot)$ is the loss function (differentiable at least once) which calculates the loss between the model prediction and ground truth, $\boldsymbol{Y}$.
$\alpha$ is a small value of $\epsilon$ that decides the size of the step to be taken in the direction of the gradient of the loss w.r.t. the input image, which leads to the input sample being perturbed such that the loss increases.
$\boldsymbol{X}^{\mathrm{adv}}$ is the adversarial sample obtained after perturbing $\boldsymbol{X}^{\mathrm{clean}}$.
To make sure that the perturbed sample is semantically indistinguishable from the unperturbed clean sample to the human eye, steps from \cref{eqn:fgsm_attack_2} and \cref{eqn:fgsm_attack_3} are performed.
Here, function $\phi^{\epsilon}$ is clipping the $\delta$ in $\epsilon$-ball for $\ell_{\infty}$-norm bounded attacks or the $\epsilon$-projection in other $l_{p}$-norm bounded attacks, complying with the $\ell_\infty$-norm or other $l_p$-norm constraints, respectively.
While function $\phi^{r}$ clips the perturbed sample ensuring that it is still within the valid input space.
FGSM, as proposed, is a single step attack.
For targeted attacks, $\boldsymbol{Y}$ is the target and $\alpha$ is multiplied by -1 so that a step is taken to minimize the loss between the model's prediction and the target prediction.

\noindent\paragraph{BIM. }This is the direct extension of FGSM into an iterative attack method. 
In FGSM, $\boldsymbol{X}^{\mathrm{clean}}$ was perturbed just once. 
While in BIM, $\boldsymbol{X}^{\mathrm{clean}}$ is perturbed iteratively for time steps $t \in [0, \boldsymbol{T}]$, such that $t\in \mathbb{Z}^+$, where $\boldsymbol{T}$ are the total number of permissible attack iterations.
This changes the steps of the attack from FGSM to the following, 
\begin{equation}
    \label{eqn:bim_attack_1}
    \boldsymbol{X}^{\mathrm{adv}_{t+1}} = \boldsymbol{X}^{\mathrm{adv}_t}+\alpha \cdot \mathrm{sign}\nabla_{\boldsymbol{X}^{\mathrm{adv}_t}}L(f_{\theta}(\boldsymbol{X}^{\mathrm{adv}_t}), \boldsymbol{Y}),
\end{equation}
%\begin{align}
%    \label{eqn:bim_attack_1}
%    \boldsymbol{X}^{\mathrm{adv}_{t+1}} &= \boldsymbol{X}^{\mathrm{adv}_t}+\alpha \cdot \mathrm{sign}\nabla_{\boldsymbol{X}^{\mathrm{adv}_t}}L(f_{\theta}(\boldsymbol{X}^{\mathrm{adv}_t}), \boldsymbol{Y}), \\
%    \boldsymbol{X}^{\mathrm{adv}_{t+1}} &= \phi^{r}(\boldsymbol{X}^{\mathrm{clean}}+ \phi^{\epsilon}(\boldsymbol{X}^{\mathrm{adv}_{t+1}} - %\boldsymbol{X}^{\mathrm{clean}})).
%\end{align}
\begin{equation}
\label{eqn:bim_attack_2}
    \delta = \phi^{\epsilon}(\boldsymbol{X}^{\mathrm{adv}_{t+1}} - \boldsymbol{X}^{\mathrm{clean}}), 
\end{equation}
\begin{equation}
\label{eqn:bim_attack_3}
    \boldsymbol{X}^{\mathrm{adv}_{t+1}} = \phi^{r}(\boldsymbol{X}^{\mathrm{clean}}+ \delta).
\end{equation}
Here, at $t$=0, $\boldsymbol{X}^{\mathrm{adv}_t}$=$\boldsymbol{X}^{\mathrm{clean}}$.

\noindent\paragraph{PGD. }Since in BIM, the initial prediction always started from $\boldsymbol{X}^{\mathrm{clean}}$, the attack required a significant amount of steps to optimize the adversarial noise and yet it was not guaranteed that in the permissible $\epsilon$-bound, $\boldsymbol{X}^{\mathrm{adv}_{t+1}}$ was far from $\boldsymbol{X}^{\mathrm{clean}}$.
Thus, PGD proposed introducing stochasticity to ensure random starting points for attack optimization.
They achieved this by perturbing $\boldsymbol{X}^{\mathrm{clean}}$ with $\mathcal{U}(-\epsilon, \epsilon)$, a uniform distribution in $[-\epsilon, \epsilon]$, before making the first prediction, such that, at $t$=0
\begin{equation}
    \label{eqn:pgd_random_start}
    \boldsymbol{X}^{{adv}_t} = \phi^{r}(\boldsymbol{X}^{clean} + \mathcal{U}(-\epsilon, \epsilon)).
\end{equation}

\noindent\paragraph{APGD. }Auto-PGD is an effective extension to the PGD attack that effectively scales the step size $\alpha$ over attack iterations considering the compute budget and the success rate of the attack.

\noindent\paragraph{CosPGD. }All previously discussed attacks were proposed for the image classification task. 
Here, the input sample is a 2D image of resolution $\mathrm{H}\times\mathrm{W}$, where $\mathrm{H}$ and $\mathrm{W}$ are the height and width of the spatial resolution of the sample, respectively.
Pixel-wise information is inconsequential for image classification.
This led to the pixel-wise loss $\mathcal{L}(\cdot)$ being aggregated to $\mathrm{L}(\cdot)$, as follows,
\begin{equation}
    \label{eqn:pixel_wise_loss}
    L(f_{\theta}(\boldsymbol{X}^{\mathrm{adv}_t}), \boldsymbol{Y}) = \frac{1}{\mathrm{H}\times\mathrm{W}}\sum_{i\in {\mathrm{H}\times\mathrm{W}}} \mathcal{L}(f_{\theta}(\boldsymbol{X}^{\mathrm{adv}_t})_i, \boldsymbol{Y}_i).
\end{equation}
This aggregation of $\mathcal{L}(\cdot)$ fails to account for pixel-wise information available in tasks other than image classification, such as pixel-wise prediction tasks like Optical Flow estimation, and disparity estimation.
Thus, in their work \cite{agnihotri2023cospgd} propose an effective extension of the PGD attack that takes pixel-wise information into account by scaling $\mathcal{L}(\cdot)$ by the alignment between the distribution of the predictions and the distributions of $\boldsymbol{Y}$ before aggregating leading to a better-optimized attack, modifying \cref{eqn:bim_attack_1} as follows,
\begin{equation}    
    \label{eqn:cospgd_attack}
\boldsymbol{X}^{\mathrm{adv}_{t+1}}=\boldsymbol{X}^{\mathrm{adv}_t}+\alpha \cdot \mathrm{sign}\nabla_{\boldsymbol{X}^{\mathrm{adv}_t}}
 \sum_{i\in H\times W}\mathrm{cos}\left(\psi(f_\theta(\boldsymbol{X}^{\mathrm{adv}_t})_i), \Psi(\boldsymbol{Y}_i)\right) \cdot \mathcal{L}\left(f_{\theta}(\boldsymbol{X}^{\mathrm{adv}_t})_i, \boldsymbol{Y}_i\right).
\end{equation}
Where, functions $\psi$ and $\Psi$ are used to obtain the distribution over the predictions and $\boldsymbol{Y}_i$, respectively, and the function $\mathrm{cos}$ calculates the cosine similarity between the two distributions.
CosPGD is the unified SotA adversarial attack for pixel-wise prediction tasks.

In \Cref{fig:adv_exp_kitti_sttr}, we show examples of adversarial attacks, on STTR using the KITTI2015 dataset. We show the samples before and after the attacks and the predictions before and after the respective adversarial attacks.

\section{Model Zoo}
\label{sec:appendix:model_zoo}

It is challenging to find all checkpoints whereas training them is time and compute-exhaustive.
Thus we gather available model checkpoints made available online by the respective authors.
The trained checkpoints for all models available in \benchmark{} can be obtained using the following lines of code:
\begin{minted}[fontsize=\small]{python}
from dispbench.evals import load_model
model = load_model(model_name='STTR', 
    dataset='KITTI2015')
\end{minted}
Each model checkpoint can be retrieved with the pair of `model\_name', the name of the model, and `dataset', the dataset for which the checkpoint was last finetuned.

\section{\benchmark{} Usage Details}
\label{sec:appendix:evaluation_details}
Following we provide a detailed description of the evaluation functions and their arguments provided in \benchmark{}.

\subsection{Adversarial Attacks}
\label{subsec:appendix:evaluation_details:adv_attack}
To evaluate a model for a given dataset, on an attack, the following lines of code are required.
\begin{minted}[fontsize=\small, breaklines]{python}
from dispcbench.evals import evaluate
model, results = evaluate( 
 model_name='STTR', dataset='KITTI2015' retrieve_existing=True,
 threat_config='config.yml')
\end{minted}
Here, the `config.yml' contains the configuration for the threat model, for example, when the threat model is a PGD attack, `config.yml' could contain `threat\_model=\textit{``PGD''}', `iterations=\textit{20}', `alpha=\textit{0.01}', `epsilon=\textit{8}', and `lp\_norm=\textit{``Linf''}'.
The argument description is as follows:
\begin{itemize}    
\item `model\_name' is the name of the disparity estimation method to be used, given as a string.
\item `dataset' is the name of the dataset to be used also given as a string. 
\item `retrieve\_existing' is a boolean flag, which when set to `True' will retrieve the evaluation from the benchmark if the queried evaluation exists in the benchmark provided by this work, else \benchmark{} will perform the evaluation.
If the `retrieve\_existing' boolean flag is set to `False' then \benchmark{} will perform the evaluation even if the queried evaluation exists in the provided benchmark.
\item The `config.yml' contains the following:
\begin{itemize}
    \item `threat\_model' is the name of the adversarial attack to be used, given as a string.
    \item `iterations' are the number of attack iterations, given as an integer.
    \item `epsilon' is the permissible perturbation budget $\epsilon$ given a floating point (float).
    \item `alpha' is the step size of the attack, $\alpha$, given as a floating point (float).
    \item `lp\_norm' is the Lipschitz continuity norm ($l_p$-norm) to be used for bounding the perturbation, possible options are `Linf' and `L2' given as a string.
    \item `target' is false by default, but to do targeted attacks, either the user can set `target'=True, to use the default target of $\overrightarrow{0}$, or can pass a specific tensor to be used as the target.
\end{itemize}
\end{itemize}

\subsection{2D Common Corruptions}
\label{subsec:appendix:evaluation_details:2dcc}
To evaluate a model for a given dataset, with 2D Common Corruptions, the following lines of code are required.
\begin{minted}[fontsize=\small, breaklines]{python}
from dispbench.evals import evaluate
model, results = evaluate( 
 model_name='STTR', dataset='KITTI2015', retrieve_existing=True,
 threat_config='config.yml')
\end{minted}
Here, the `config.yml' contains the configuration for the threat model; for example, when the threat model is 2D Common Corruption, `config.yml' could contain `threat\_model=\textit{``2DCommonCorruption''}', and  `severity=\textit{3}'.
Please note, when the `threat\_model' is a common corruption type, \benchmark{} performs evaluations on all corruptions under the respective `threat\_model' and returns the method's performance on each corruption at the requested severity.
The argument description is as follows:
\begin{itemize}    
\item `model\_name' is the name of the disparity estimation method to be used, given as a string.
\item `dataset' is the name of the dataset to be used also given as a string. 
\item `retrieve\_existing' is a boolean flag, which when set to `True' will retrieve the evaluation from the benchmark if the queried evaluation exists in the benchmark provided by this work, else \benchmark{} will perform the evaluation.
If the `retrieve\_existing' boolean flag is set to `False', then \benchmark{} will perform the evaluation even if the queried evaluation exists in the provided benchmark.
\item The `config.yml' contains the following:
\begin{itemize}
    \item `threat\_model' is the name of the common corruption to be used, given as a string, i.e.~`2DCommonCorruption'.
    \item `severity' is the severity of the corruption, given as an integer between 1 and 5 (both inclusive).
\end{itemize}
\end{itemize}
\benchmark{} supports the following 2D Common Corruption: `gaussian\_noise', shot\_noise', `impulse\_noise', `defocus\_blur', `frosted\_glass\_blur', `motion\_blur', `zoom\_blur', `snow', `frost', `fog', `brightness', `contrast', `elastic', `pixelate', `jpeg'.
For the evaluation, \benchmark{} will evaluate the model on the validation images from the respective dataset corrupted using each of the aforementioned corruptions for the given severity, and then report the mean performance over all of them.

\section{Extension To Analysis}
\label{sec:appendix:analyis_extension}
Following, we extend the analysis from \Cref{subsec:analysis:findings} and report additional evaluations from \benchmark{}.

\subsection{KITTI2015}
\label{sec:appendix:analysis_extension:kitti}
Following, we provide evaluations of on the KITTI2015 dataset.
\begin{figure}[h]
    \centering
    \includegraphics[width=1.0\linewidth]{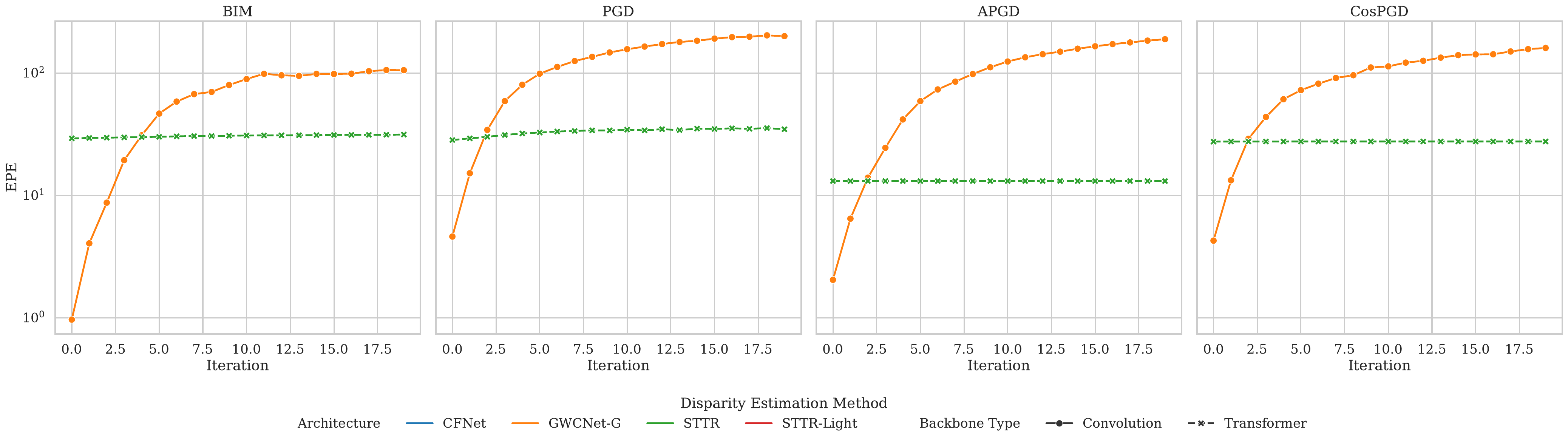}
    \caption{Evaluations of all considered adversarial attacks with $\epsilon=\frac{8}{255}$ and $\alpha$=0.01 under the $\ell_{\infty}$-norm bound using the KITTI2015 validation dataset.}
    \label{fig:adv_attack_kitti}
\end{figure}
In \Cref{fig:adv_attack_kitti} we report the evaluations of all considered adversarial attacks with $\epsilon=\frac{8}{255}$ and $\alpha$=0.01 under the $\ell_{\infty}$-norm bound using the KITTI2015 validation dataset. 

\begin{figure*}
    \centering
    \rotatebox{0}{
    \includegraphics[width=0.55\linewidth]{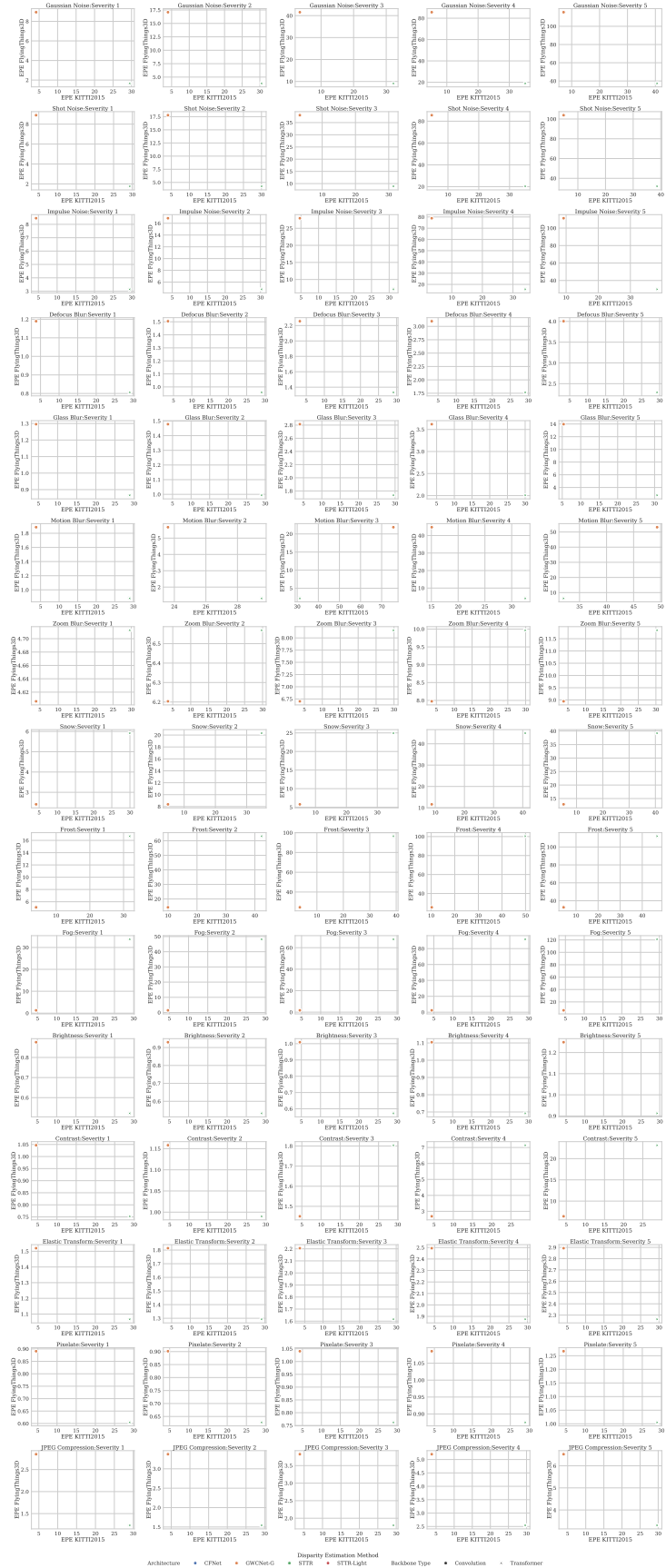}
    }
    \caption{For the same architecture, we evaluate checkpoints pretrained on Flyingthings3D against synthetic 2D Common Corruption on Flyingthings3D and correlate its performance with the checkpoint trained on KITTI2015 against synthetic 2D Common Corruptions on KITTI2015.
    Here, we report the EPE across every individual corruption for a given severity level. 
    We observe no correlation in performance, indicating that synthetic corruptions on synthetic datasets cannot be used as a proxy for real-world corruptions.}
    \label{fig:all_correlation_synthetic_real}
\end{figure*}

In \Cref{fig:all_correlation_synthetic_real}, we extend the evaluations from \Cref{fig:mean_correlation_synthetic_real}, showing that the observations made over the mean performance over all corruptions also hold for every individual corruption.

\iffalse
\section*{Limitations}
%\label{subsec:conclusion:limitations}
%\noindent\paragraph{Limitations. }
Benchmarking disparity estimation methods is a compute and labor-intensive endeavor. 
Thus, best utilizing available resources, we currently use \benchmark{} to benchmark a very limited number of settings, using the most popular works for now.
The benchmarking tool itself offers significantly more combinations that can be benchmarked.
\fi

\end{document}